\documentclass[12pt]{article}

\usepackage{newtxtext,newtxmath}

\usepackage{graphicx}
\usepackage{booktabs}
\usepackage{caption}
\usepackage{booktabs}
\usepackage{multirow}

\usepackage[table]{xcolor}

\usepackage[letterpaper,margin=1in]{geometry}

\renewenvironment{abstract}
	{\quotation}
	{\endquotation}

\date{}

\makeatletter
\renewcommand{\fnum@figure}{\textbf{Figure \thefigure}}
\renewcommand{\fnum@table}{\textbf{Table \thetable}}
\makeatother

\usepackage{scicite}

\usepackage{url}

\def\scititle{
Long-Short Term Agents for Pure-Vision Bronchoscopy Robotic Autonomy
}
\title{\bfseries \boldmath \scititle}

\author{
	Junyang Wu$^{1}$,
	Mingyi Luo$^{1}$,
    Fangfang Xie$^{2}$,
    Minghui Zhang$^{1}$,
    Hanxiao Zhang$^{1}$,\\
    Chunxi Zhang$^{2}$,
    Junhao Wang$^{1}$,
    Jiayuan Sun$^{2\ast}$,
    Yun Gu$^{2\ast}$,
    Guang-Zhong Yang$^{2\ast}$\and
	\small$^{1}$Institute of Medical Robotics, Shanghai Jiao Tong University, Shanghai, 200240, China.\and
	\small$^{2}$Shanghai Chest Hospital, Shanghai, 10587, China.\and
	\small$^\ast$Corresponding author.
}

\begin{document} 

\maketitle

\begin{abstract} \bfseries \boldmath
Accurate intraoperative navigation is essential for robot-assisted endoluminal intervention, but remains difficult because of limited endoscopic field of view and dynamic artifacts. Existing navigation platforms often rely on external localization technologies, such as electromagnetic tracking or shape sensing, which increase hardware complexity and remain vulnerable to intraoperative anatomical mismatch.
We present a vision-only autonomy framework that performs long-horizon bronchoscopic navigation using preoperative CT-derived virtual targets and live endoscopic video, without external tracking during navigation. The framework uses hierarchical long-short agents: a short-term reactive agent for continuous low-latency motion control, and a long-term strategic agent for decision support at anatomically ambiguous points. When their recommendations conflict, a world-model critic predicts future visual states for candidate actions and selects the action whose predicted state best matches the target view.
We evaluated the system in a high-fidelity airway phantom, three ex vivo porcine lungs, and a live porcine model. The system reached all planned segmental targets in the phantom, maintained 80\% success to the eighth generation ex vivo, and achieved in vivo navigation performance comparable to the expert bronchoscopist. These results support the preclinical feasibility of sensor-free autonomous bronchoscopic navigation.

\end{abstract}

\section*{Summary}
We present a pure-vision imitation-learning framework achieving long-horizon autonomy in robotic bronchoscopy navigation.

\section{Introduction}

Endoluminal intervention has emerged as a paradigm-shifting modality for the integrated diagnosis and treatment of early-stage luminal cancers. By utilizing natural orifices, these procedures offer a minimally invasive alternative to traditional resection for digestive, pulmonary, urinary, and gynecologic pathologies\cite{yamamoto2007technology, mannath2016role, ngamruengphong2025aga, pan2014endoscopic, deng2023efficacy}. The recent introduction of endoluminal robotic platforms has further augmented this capability, enhancing the safety, maneuverability, and precision of instrument manipulation\cite{martin2020enabling, hindson2020intelligent, mao2024magnetic, xiao2021fully, norton2019intelligent}. However, accurate navigation within deep, tortuous, and unstructured anatomical environments remains a fundamental bottleneck. This challenge is exacerbated by the constraints of the endoluminal environment: a limited endoscopic field-of-view (FoV), significant \textit{in vivo} artifacts such as fluid occlusion and motion blur, and the absence of distinctive geometric landmarks in deformable tissue \cite{ali2022we, kennedy2020computer, lu2023assessment, wu2025trimming, schmidt2024tracking, maas2024computer}.

Current clinical navigation systems rely primarily on external localization technologies such as electromagnetic tracking and shape sensing\cite{gildea2006electromagnetic, hautmann2005electromagnetic, folch2020sensitivity, makris2007electromagnetic, kalchiem2022shape, husta2025shape, eberhardt2007electromagnetic}. These platforms can improve procedural guidance, but they also introduce additional hardware, workflow complexity, and cost. More importantly, they depend on registration between preoperative imaging and intraoperative anatomy. In bronchoscopy, this registration is vulnerable to respiratory motion, tissue deformation, and instrument interaction, which can produce clinically important CT-to-body divergence\cite{chan2024new, paez2025robotic, pritchett2020virtual}. Electromagnetic sensing is also susceptible to interference from metallic instruments\cite{franz2014electromagnetic, sorriento2019optical, stevens2010minimizing}. These limitations motivate navigation strategies that rely more directly on the intraoperative scene.

Pure vision-based navigation offers an attractive alternative because the endoscope already provides continuous image feedback at the site of intervention. In principle, this could reduce dependence on external tracking hardware while improving adaptation to local anatomy and real-time procedural changes. However, achieving robust control from endoscopic images alone remains difficult. 
Although modern computer vision methods have substantially advanced endoscopic perception, including diagnostic support, depth estimation, and scene understanding\cite{cao2023intelligent, igaki2023automatic, wang2018development, zhou2020diagnostic, fockens2023deep, luo2019real, chen2025artificial, shao2022self}, most of these systems are designed to understand the scene rather than to close the control loop for long-horizon autonomous navigation.

Recent work in surgical robotics has begun to demonstrate higher levels \cite{yang2017medical} of autonomy across specialized tasks.
In the domain of ophthalmic microsurgery, where physiological tremors limit human performance, Bian et al.\cite{bian2026autonomous} developed the ARISE system, which leverages multiview spatial fusion to resolve depth perception deficits for autonomous subretinal injections, while Zhang et al. \cite{zhang2025deep} utilized deep learning frameworks to automate retinal vein cannulation, achieving high success rates even under respiratory-induced motion. Addressing the challenges of abdominal soft-tissue manipulation, Kim et al. \cite{kim2025srt} introduced the Hierarchical Surgical Robot Transformer (SRT-H), a framework that decouples high-level language planning from low-level motor policies to enable complex decision-making and error recovery in cholecystectomy; concurrently, Long et al. \cite{long2025surgical} demonstrated a surgical embodied intelligence approach capable of zero-shot sim-to-real transfer, facilitating generalized task execution in unstructured laparoscopic settings. Furthermore, in pulmonary interventions, Liu et al. \cite{liu2025ai} proposed a memory-augmented robotic bronchoscope that autonomously navigates the bronchial tree for foreign body retrieval without preoperative imaging, whereas Kuntz et al. \cite{kuntz2023autonomous} presented a steerable needle system that integrates respiratory gating and motion-compensated replanning to precisely biopsy peripheral lung nodules. 
While recent works demonstrate efficacy in executing discrete surgical sub-tasks, they lack the capacity to support comprehensive, end-to-end procedural autonomy.

In bronchoscopy specifically, prior navigation methods have shown promising short-horizon control through explicit modeling, including lumen-centered visual servoing, depth-guided steering, and centerline-constrained policies trained within the virtual engine. 
However, these approaches often depend on hand-crafted heuristics, local geometric cues, or static anatomical assumptions, which limits robustness over long trajectories in the highly repetitive and deformable bronchial tree.
We therefore pursue an imitation-learning paradigm, in which an agent learns directly from expert demonstrations to capture the sequential decision structure used by skilled bronchoscopists, with the goal of approaching human-level performance in clinically realistic navigation tasks.

In this paper, we present a vision-only autonomous bronchoscopy system that navigates using endoscopic video and preoperative CT, without external localization hardware. 
The framework combines a short-term reactive controller for continuous maneuvering, a long-term strategic module for decision support at anatomically ambiguous points, and a world-model critic for resolving conflicting actions. 
We evaluate the system in a high-fidelity phantom, ex vivo porcine lungs, and a live porcine model under active respiration. Across these progressively realistic settings, the system achieved reliable distal airway access, with endpoint accuracy approaching that of expert operators.

\section{Results}

In this section, we first outline the vision-only navigation workflow. We then evaluate its performance across a spectrum of increasingly realistic environments: a high-fidelity phantom with full lung segment reach, controlled visual perturbation tests, diverse ex vivo porcine lungs, and a live porcine model under active respiration. The primary outcomes assessed include navigation success rate, time to target, and the total number of actions executed.

\begin{figure}[]
\centering
\includegraphics[width=0.9\linewidth]{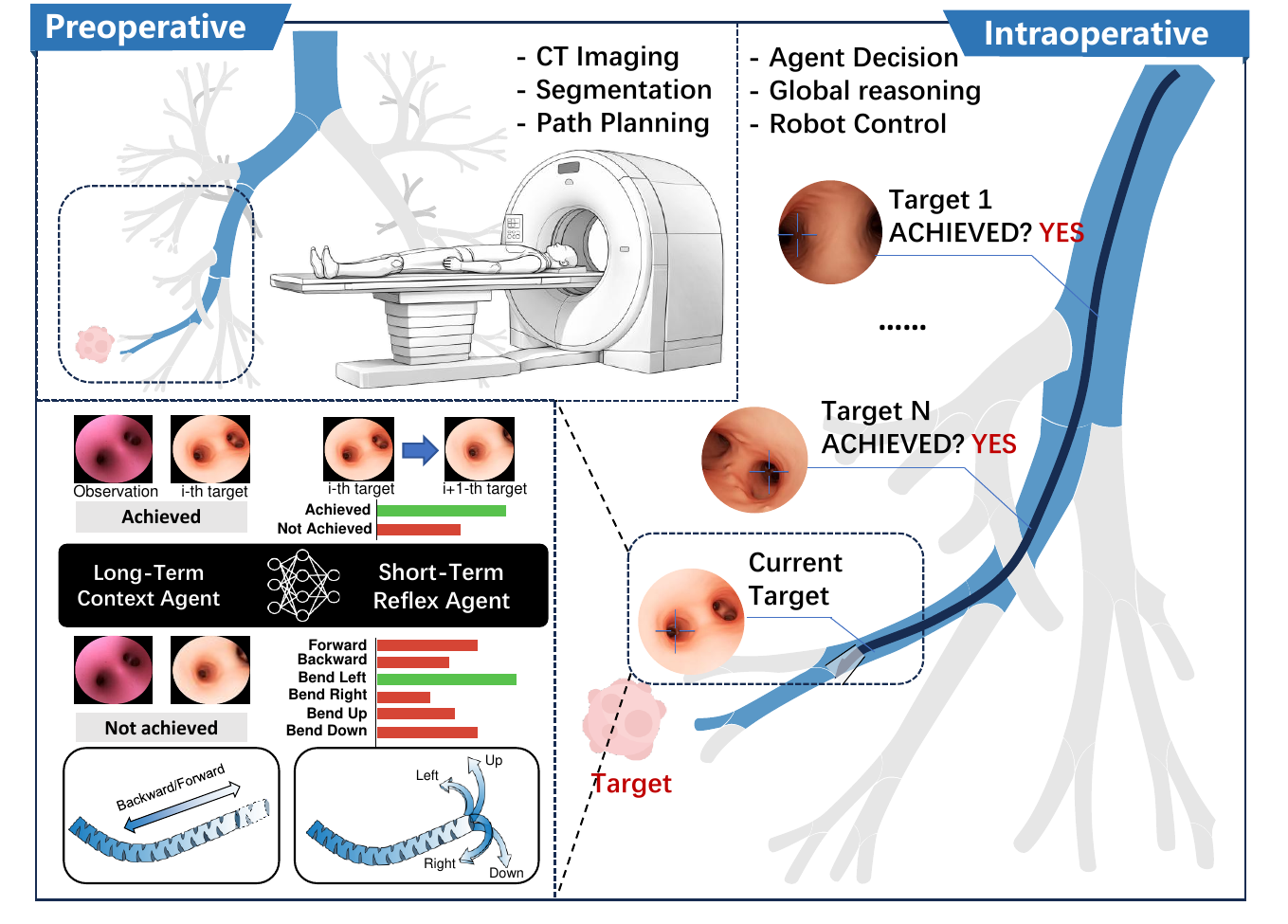}
\caption{\textbf{Conceptual overview of the autonomous robotic navigation framework.} Preoperatively, following a patient CT scan, fully automated algorithms are deployed to segment the bronchial tree and target lesions, and to plan the optimal intraoperative trajectory. This planned path is formulated as a sequential series of virtual image targets. Intraoperatively, the intelligent agent autonomously navigates through these consecutive sub-targets to ultimately access the bronchial segment nearest to the lesion. To achieve this, the agent continuously executes a dual-state decision policy: if it determines that the robot has successfully reached the current sub-target, it updates the objective to the subsequent waypoint; otherwise, it generates specific kinematic control commands to actuate the 6-degree-of-freedom (6-DoF) flexible robot toward the current target.}

\label{fig:concept_sr} 
\end{figure}

\subsection{System overview and workflow}

As illustrated in Fig. \ref{fig:concept_sr}, we present a vision-only autonomous framework for bronchoscopic navigation. The workflow begins with preoperative CT, from which we segment the airway tree \cite{zhang2024airmorph}, segment the target nodule, and plan the optimal intraoperative navigation path autonomously. 
We then render a sequence of virtual bronchoscopic views along this path, which serves as the visual target trajectory during autonomous navigation.

To make long-horizon navigation tractable, we divide the planning path into image-based sub-targets. 
During the intraoperative phase, the robot traverses through these intermediate waypoints sequentially to ultimately reach the target region. 
It is governed by a unified control policy tasked with two concurrent objectives: (i) while the active sub-target remains unreached, generating continuous kinematic commands to visually align the live endoscopic view to the current virtual target; and (ii) continuously evaluating the visual alignment status to determine sub-target switch. Upon confirming arrival, the system dynamically advances the next sub-target. In this way, navigation is performed as a sequential visual matching process.

\begin{figure}[]
\centering
\includegraphics[width=0.95\linewidth]{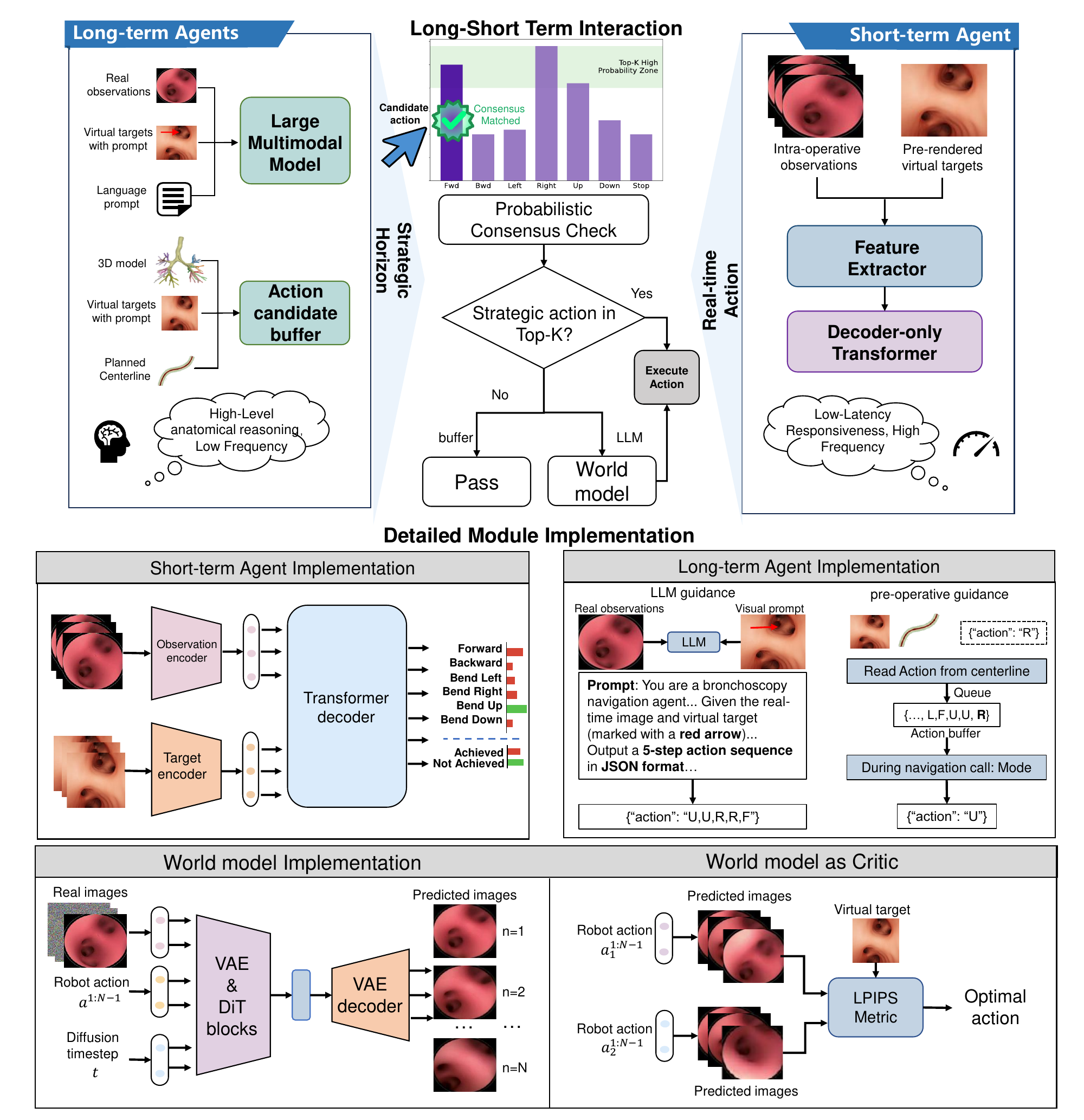}
\caption{\textbf{Architecture of the hierarchical multi-agent autonomous navigation framework.} The system decomposes the navigation task across two temporal scales: a short-term reactivate agent that compensates for immediate endoluminal dynamics, and a long-term strategic agent that acts as a high-level supervisor. Inter-agent coordination is governed by an interaction and consensus mechanism. If the strategic agent’s proposed action aligns with the top-K predicted logits of the reactive agent, a consensus is achieved and the action is executed. In the event of a conflict, resolution is source-dependent: preoperative guidance commands will not be executed. Conversely, if a conflict originates from the LLM guidance, a world model serves as a critic to simulate potential downstream outcomes and deduce the optimal control signal.}

\label{fig:world_model_vis} 
\end{figure}

\subsubsection{Intra-operative multi-agents design}

Our autonomous navigation framework is structured as a hierarchical multi-agent system, comprising a \textbf{Short-term Reactive Agent} and a \textbf{Long-term Strategic Agent}.  
The short-term reactivate agent provides continuous, high-frequency command execution for the vast majority of the intraoperative procedure, adapting instantaneously to real-time endoluminal dynamics. 
The long-term strategic agent is sparsely invoked—triggered only at critical anatomical decision points, such as bronchial bifurcations, or by specific procedural anomalies. When queried, this Strategic Agent aggregates two complementary guidance modalities: \textit{Pre-operative Guidance}, derived from the static anatomical roadmap, and \textit{LLM Guidance}, powered by a large multimodal model for high-level semantic reasoning.

To coordinate these agents, we implement a robust Interaction and Consensus Mechanism. When both agents are activated simultaneously, the system evaluates the consistency of their proposed actions. Specifically, if the action proposed by the Long-term Agent falls within the top-K predicted logits of the Short-term Reactive Agent, a consensus is established, and the action is executed immediately. If this condition is not met, a conflict is identified. 
In such cases, the resolution logic depends on the source of the strategic command. 
If the action originates from the Pre-operative Guidance, it will not be executed. Conversely, if the conflicting action originates from the LLM Guidance, the system invokes the world model to simulate potential outcomes and identify the optimal control signal.

\paragraph{Short-term Reactive Agent Design}
To ensure low-latency responsiveness, the Short-term Reactive Agent is built upon a lightweight Transformer architecture. 
An EfficientNet-B0 \cite{koonce2021efficientnet} backbone encodes the current endoscopic frame and the active virtual target into compact visual features, which are concatenated and fed into a decoder-only Transformer.
The action space includes forward and backward translation, four directional bending commands, and a target-switch command indicating that the current waypoint has been reached. 
The decision to transition to the next virtual target is explicitly incorporated as an intrinsic component of the action space, enabling the agent to autonomously determine the optimal timing for reference switching. 
The agent is trained by imitation learning with cross-entropy loss on expert demonstrations.

\paragraph{Long-term Strategic Agent Design}
The long-term strategic agent combines two guidance sources. 
The \textit{Pre-operative Guidance} component derives virtual trajectories from geometric centerlines extracted from preoperative CT scans. 
We log all virtual poses along the centerlines point and transferred them to the action space. Intraoperatively, these actions are buffered across the preceding 10 frames, and the dominant action is selected by majority voting.
The \textit{LLM Guidance} integrates a Large Multimodal Model to provide semantic guidance in ambiguous visual situations. 
A large multimodal model receives virtual target images annotated with a directional arrow indicating the target lumen, together with a structured text prompt. Based on these visual and textual inputs, the model proposes a short sequence of high-level actions.

\paragraph{World Model as Critic}
In scenarios of conflicting directives between the reactive and strategic agents, a world model acts as a critic. For each candidate action, the model predicts a short rollout of future endoscopic frames and compares the predicted visual state with the target virtual view using the Learned Perceptual Image Patch Similarity (LPIPS). 
By computing perceptual distances, the critic measures the semantic consistency of each future state.
The action that minimizes perceptual discrepancy is selected as the optimal control decision.

\subsection{Evaluation of Full Segmental Reach in a High-Fidelity Phantom}

We first tested whether the system could sustain long-horizon navigation across the full segmental structure of the lung using a high-fidelity airway phantom. We utilized AirwayNet \cite{zhang2024airmorph} for bronchial tree segmentation and segment-level classification, identifying 17 anatomical lung segments consistent with human anatomy (Fig. \ref{fig:phantom_eval}A). To validate global reachability, we planned 17 distinct trajectories covering all identified segments. For each trajectory, the autonomous system was initialized at the main trachea and tasked with navigating to a designated segmental target under the continuous supervision of an expert bronchoscopist. A failure was defined as entry into a bronchial segment inconsistent with the planned path; in such instances, the trial was terminated, and the preceding segment was recorded as the terminal endpoint.

Our method was benchmarked against expert manual teleoperation (Expert) and two autonomous baselines: GNM \cite{shah2023gnm} and ViNT \cite{shah2023vint}. Qualitative comparisons (Fig. \ref{fig:phantom_eval} C and Fig. \ref{fig:supp1}) demonstrate that our method reaches terminal endpoints indistinguishable from those obtained by the expert, confirming high-fidelity navigation.

Fig. \ref{fig:phantom_eval} E and H detail navigation performance across the 17 trajectories, quantifying the frequency with which each method reached specific bronchial generations. Success rates for all methods naturally declined as airway depth increased. With the exception of a single failure by ViNT, all methods successfully navigated to the second generation. 
Only the Expert and our method navigated beyond the eighth generation, with our approach matching Expert performance and reaching all generations attained by the human operator. The maximum depth was limited solely by physical constraints, when the phantom airway diameter fell below the bronchoscope diameter (4.2 mm).

Quantitative analysis reveals a distinct operational trade-off between task duration and control efficiency (Fig. \ref{fig:phantom_eval} F). While our autonomous system required more time to complete the navigation than expert teleoperation ($450.7 \pm 69.5$ s versus $273.5 \pm 77.5$ s), it required significantly fewer control actions ($275.8 \pm 31.9$ versus $346.8 \pm 45.9$; $P < 0.001$). This reduction in control actions suggests that the robot's autonomous navigation minimizes the redundant motion and micro-adjustments characteristic of manual teleoperation.
Importantly, the increase in procedure time was primarily due to a deliberately imposed 3-second safety window for each mechanical execution step (Table \ref{tab:timing_analysis}), rather than model inference latency (merely 6 ms).

We further assessed the maximum bronchial generation reached across the 17 segments (Fig. \ref{fig:phantom_eval} I). Our method reached the terminal planning target in every segment, matching the Expert's coverage. The system achieved a mean generation depth of $5.53 \pm 1.55$, significantly outperforming both GNM ($4.24 \pm 1.60$) and ViNT ($3.65 \pm 1.62$) baselines. 
To quantify visual alignment consistency, 
our method achieved a significantly higher SSIM ($0.841 \pm 0.066$) compared to the ViNT baseline ($0.776 \pm 0.044$) (Fig. \ref{fig:phantom_eval} J). Crucially, as the ViNT baseline shares the same underlying network architecture with our short-term reactive agent, the performance difference highlights the importance of the proposed hierarchical multi-agent framework. Although ViNT performs comparably in short-range bronchial segments, error accumulation over long-horizon trajectories leads to navigation failure in complex airway structures (Fig. \ref{fig:supp3}). This confirms that collaborative strategic guidance is the key mechanism enabling robust, whole-lung navigability.


\subsection{Robustness to Visual Perturbations}

\begin{figure}[]
\centering
\includegraphics[width=0.73\linewidth]{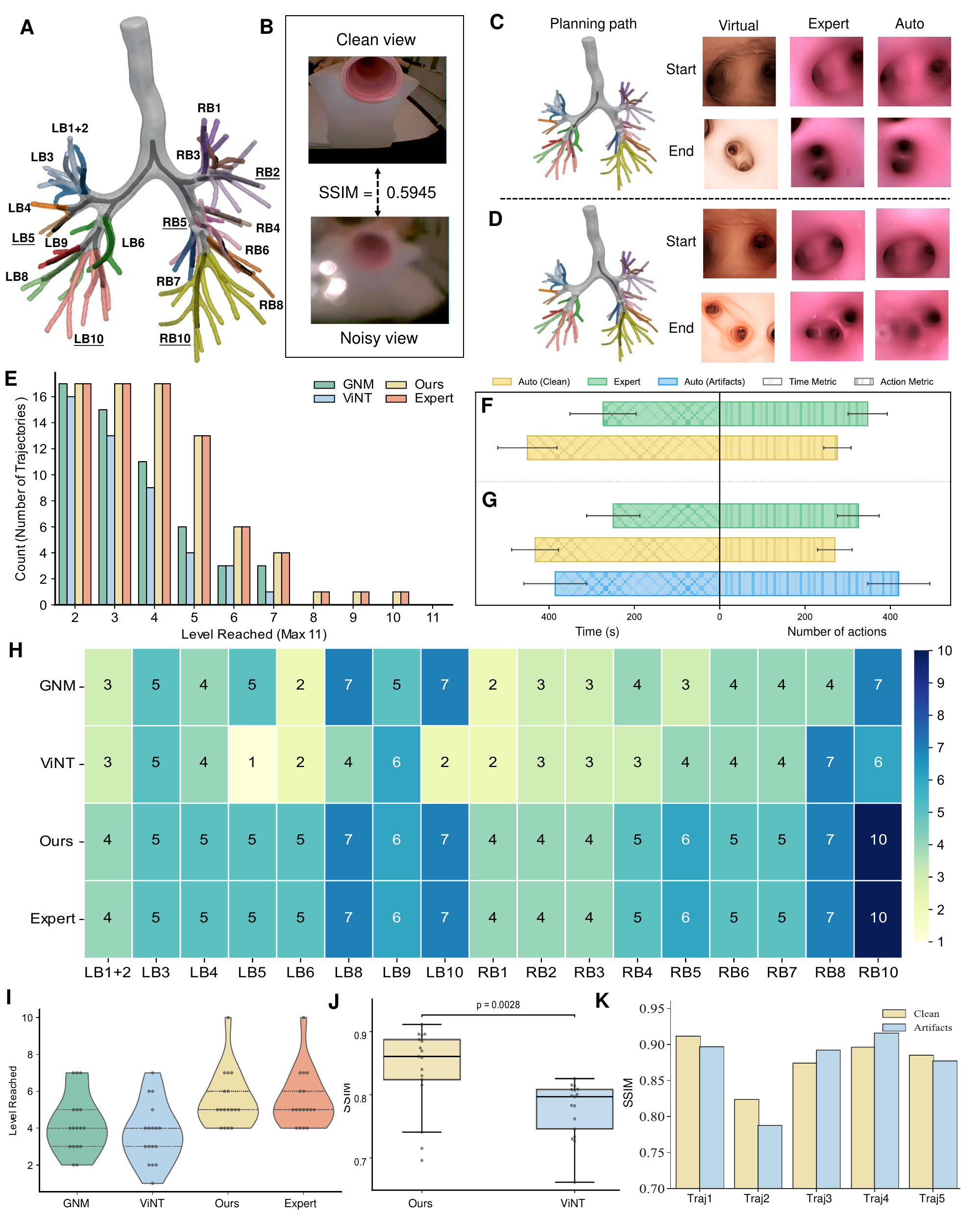}
\caption{\textbf{Results of phantom experiments.} A. Bronchial segmentation of the experimental phantom displaying 17 planned trajectories (five of which were utilized for artifact experiments). B. A comparison between artifact-degraded and clean images.
C. A successful trajectory of clean phantom.
D. A successful trajectory of artifact phantom.
E. Frequency distribution of the levels reached by different methods.
F. Comparison of time and number of actions across different methods on the clean phantom.
G. Comparison of time and number of actions across different methods on the artifact phantom.
H. Details of the maximum bronchial generation reached by different methods for each of the 17 trajectories.
I. Quantitative analysis of the bronchial generations reached by different methods.
J. SSIM between the expert's view and the final images obtained by our method and ViNT upon reaching the endpoint.
K. SSIM comparison between the final images of our method and the expert's view in both clean and artifact phantom scenarios.}
\label{fig:phantom_eval} 
\end{figure}

To test robustness under clinically relevant visual degradation, we introduced controlled lens contamination by applying glycerol to the endoscopic camera. It introduced significant optical distortion and reduced the baseline Structural Similarity Index (SSIM) between clear and soiled views to 0.59 (Fig. \ref{fig:phantom_eval} B). Five trajectories spanning different lung regions (targeting LB5, LB10, RB2, RB5, and RB10) were then executed under these challenging conditions.

Qualitative results (Fig. \ref{fig:phantom_eval} D) demonstrate that even under strong visual perturbations, the autonomous agent successfully navigated to designated targets, tracing paths aligned closely with expert trajectories. However, quantitative analysis reveals a distinct trade-off between robustness and efficiency (Fig. \ref{fig:phantom_eval} G). 
In particular, the number of control actions increases substantially under artifact conditions ($419.4 \pm 72.9$) compared with clean autonomous navigation ($270.0 \pm 40.4$) (Fig. \ref{fig:phantom_eval} G). This elevated action count reflects frequent re-planning and fine-grained corrective maneuvers, revealing increased hesitation and back-and-forth adjustments as the agent actively compensates for perceptual uncertainty induced by visual degradation.
In terms of temporal metrics, the total execution time under artifact conditions was $385.5 \pm 73.6$ s, compared with $249.5 \pm 62.3$ s for expert teleoperation. Notably, this duration was slightly shorter than that of autonomous navigation in the clean phantom setting ($432.7 \pm 54.9$ s). Post hoc review of the experiment videos (Supp Video S3) indicated that the additional time observed in the clean phantom trials was primarily associated with longer LLM-guidance calls and greater network latency.

Under degraded imaging, the system successfully completed four of the five predefined trajectories, with failure observed only in the anatomically complex right upper lobe (RB2). To assess endpoint fidelity, we compared the visual state at the agent’s terminal position with the corresponding expert-defined target view (Fig. \ref{fig:phantom_eval} K). The SSIM under artifact conditions ($0.874 \pm 0.051$) was statistically indistinguishable from that achieved in the clean setting ($0.878 \pm 0.033$). These results indicate that, although visual artifacts introduce procedural delays, they do not impede convergence to the correct anatomical endpoint, demonstrating the robustness of our system.

\begin{figure}[]
\centering
\includegraphics[width=0.9\linewidth]{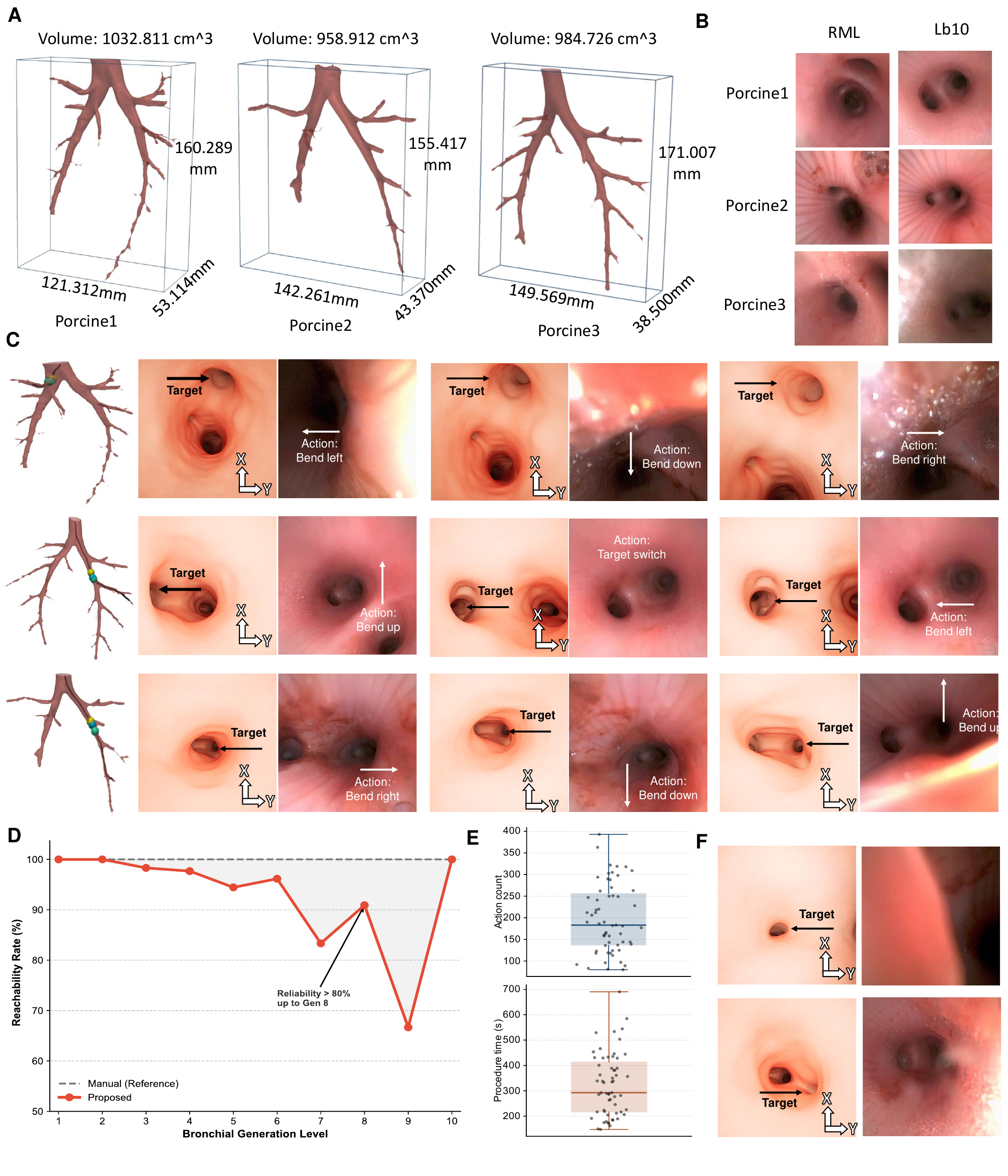}
\caption{\textbf{Ex vivo evaluation on diverse porcine lungs.} (A) 3D segmentation of three distinct porcine lungs showing morphological variability. (B) Endoscopic views at identical anatomical landmarks across different lungs, highlighting visual domain gaps. (C) System response to specific challenges: navigating through mucus occlusion, autonomous target switching upon visual matching, and adaptive view adjustment to avoid obstructions. (D) Success rates of 59 navigation trajectories across bronchial generations. (E) Distribution of procedure time and action steps. (F) Failure modes caused by lens fouling and complete bubble occlusion.}
\label{fig:exvivo_porcine} 
\end{figure}

\subsection{Ex Vivo Evaluation of Bronchial Navigation}

To assess generalization beyond rigid phantoms, we evaluated the system in three ex vivo porcine lungs spanning different airway morphologies (Fig. \ref{fig:exvivo_porcine}A). These specimens introduce significant unstructured variability, including distinct airway morphologies, non-rigid tissue deformation, and the presence of secretions such as mucus, blood, and bubbles. As illustrated in Fig. \ref{fig:exvivo_porcine}B, this variability creates a substantial perceptual challenge: endoscopic views at identical anatomical landmarks exhibit drastic textural and lighting differences across subjects, rigorously testing the system's visual generalization.

Fig. \ref{fig:exvivo_porcine}C demonstrates the system's robustness within these unstructured environments.
In the first case, where the target lumen was partially occluded by mucus, the agent exhibited resilience to visual noise by maintaining a correct \textit{rightward bending action} aligned with the anatomical target. In the second case, upon achieving high perceptual similarity to the virtual reference, the system autonomously triggered the \textit{target-switching}, effectively replicating expert sequential decision-making. In the third case, where mucus obscured the lower-right field of view, the system executed an \textit{upward bending action} to regain a clear line of sight. These adaptive responses indicate that through large-scale imitation learning, the agent has encoded complex recovery strategies analogous to the compensatory techniques used in clinical bronchoscopy. More examples are shown in Fig. \ref{fig:supp4}.

Quantitative performance across 59 distinct navigation trajectories is summarized in Fig. \ref{fig:exvivo_porcine}D. The system maintained a success rate exceeding 80\% for targets located up to the eighth bronchial generation, confirming reliable deep-lung access. Operational metrics (Fig. \ref{fig:exvivo_porcine}E) yielded a mean procedure time of $323.82 \pm 123.92$ s and a mean control action count of $198.12 \pm 79.34$ per trajectory. Despite high overall success, specific failure modes were identified (Fig. \ref{fig:exvivo_porcine}F). Navigation failed primarily when persistent artifacts (e.g., mucus) adhered directly to the camera lens, permanently obstructing the visual field, or when bubbles completely occluded the target lumen, rendering the anatomical destination invisible. These cases delineate the current physical boundaries of visual navigation under severe environmental occlusion.


\subsection{Autonomous navigation in in vivo porcine model}

\begin{figure}[]
\centering
\includegraphics[width=0.8\linewidth]{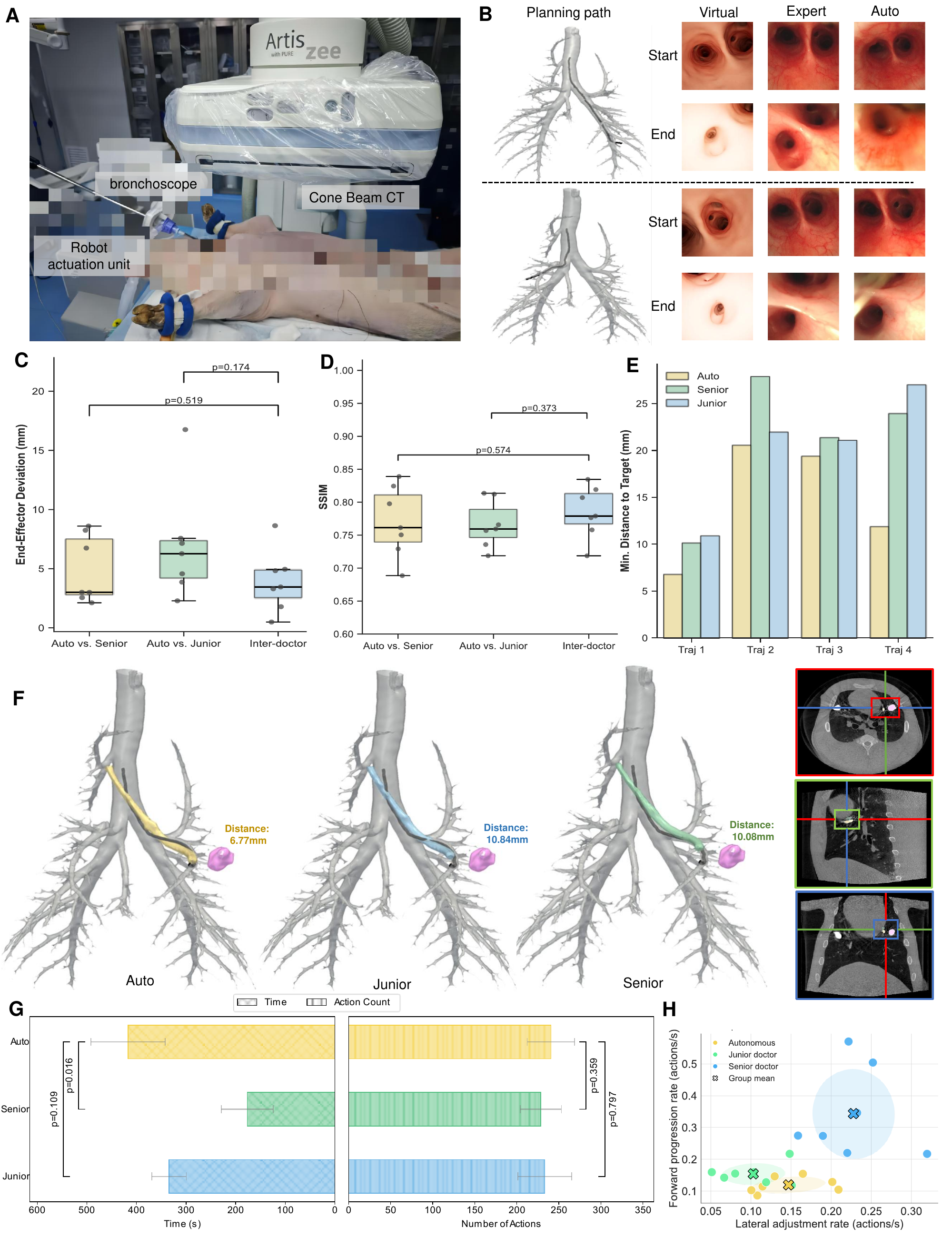}
\caption{\textbf{Results of in-vivo experiments.} A. Experimental setup for in vivo studies. B. Two examples by the automated system and the human expert. C. End-effector distance between the expert and the automated system. D. Structural Similarity Index Measure (SSIM) of the endpoints between the expert and the automated system. E. Distance to the nodule upon reaching the endpoint across different methods. F. Intraoperative CBCT images displaying preoperative bronchial and nodule segmentations, as well as intraoperative bronchoscope segmentation. G. Quantitative analysis of time and number of actions across different methods. H. Analysis of the lateral adjustment rate and the forward progression rate.
}

\label{fig:invivo} 
\end{figure}

To evaluate the clinical translational potential of our vision-based navigation system, 
we evaluated the system in a live porcine model under active respiration, which introduced airway deformation, motion, and specular visual artifacts absent in static settings.
To ensure a comprehensive and diverse validation of the system's capabilities, we defined a total of seven distinct navigation tasks traversing varying bronchial topologies. 
Four tasks targeted artificial nodules implanted into peripheral segments of the Left Upper Lobe (LUL), Left Lower Lobe (LLL), Right Upper Lobe (RUL), and Right Lower Lobe (RLL). Three additional distal targets were arbitrarily designated by the attending physician within non-nodular branches (Fig. \ref{fig:supp2}).

To benchmark the performance against the clinical gold standard, a senior bronchoscopist ($>$10 years of experience) was recruited to perform the same navigation tasks via teleoperation. We also established a clinical baseline for inter-operator variability by recording the performance of a junior bronchoscopist. For each task, the autonomous system and expert navigated from the trachea using endoscopic image feedback and the preoperative CT-derived planning path (Fig. \ref{fig:supp7}), without external localization sensors during navigation.
As shown in Fig. \ref{fig:invivo} B, starting from nearly identical initialization points, our system reached terminal positions that closely matched those of the senior expert.

We evaluated the performance of the autonomous system using two complementary metrics: spatial positioning accuracy and visual alignment consistency. 
Spatial positioning accuracy, verified via intraoperative Cone Beam CT (CBCT), measures the spatial concordance between the autonomous system’s terminal position and the target location manually navigated by the expert.
Visual alignment consistency compares the similarity between the terminal endoscopic view of the autonomous system and that of the expert, serving as a proxy for orientational and angular alignment. We posit that these two metrics are mutually reinforcing; while CBCT provides precise ground-truth spatial coordinates, it cannot capture the fine-grained angular orientation of the endoscope tip—a deficiency effectively addressed by visual alignment analysis.

First, we evaluated spatial positioning accuracy using intraoperative CBCT. For each of the seven targets, we acquired CBCT scans of the final destinations reached by a senior expert, a junior expert, and the autonomous navigation algorithm.
The autonomous system achieved a 100\% success rate (7/7 targets), consistently reaching the target. We quantified the Euclidean distance between the senior expert's and the robot's endpoints in the CBCT coordinate space. As quantified in Fig.  \ref{fig:invivo} C, the mean deviation was $4.90 \pm 2.64$ mm, which was of similar magnitude to the difference observed between the senior and junior bronchoscopists ($3.92 \pm 2.42$ mm). This indicates that, the autonomous system reached anatomically similar terminal positions to those achieved by human operators.

We next assessed visual alignment consistency using SSIM. 
As shown in Fig. \ref{fig:invivo} D, quantitative analysis across the seven trajectories yielded an average SSIM of $0.7701 \pm 0.0564$ for the autonomous system relative to the senior expert (Auto vs. Senior), which was similar to the inter-operator value between senior and junior bronchoscopists ($0.7847 \pm 0.0401$).
Together with the CBCT analysis, this indicates that the autonomous agent not only reached the correct anatomical segment but also achieved an orientation and visual perspective within the range of operator-to-operator variation observed in this study.

Furthermore, in the nodule targeting tasks (Fig. \ref{fig:invivo} E), the autonomous system reached distances to the target that were comparable to, and in some cases smaller than, those achieved by manual navigation. In all four test cases, the agent achieved minimum distances to the nodule ranging from 6.77 mm to 20.55 mm, outperforming or matching manual navigation particularly in complex trajectories (e.g., Traj 1: Auto 6.77 mm vs. Senior 10.08 mm). These results confirm that our agent can reliably guide the bronchoscope to the target region with high precision.
Fig. \ref{fig:invivo} F and Fig. \ref{fig:supp6} present the in vivo validation of trajectories using CBCT. We visualize the preoperative airway segmentation and planned centerline alongside the intraoperative segmentation of the endoscope. It is important to note that due to physiological respiratory motion in the porcine model and metallic artifacts inherent to CBCT imaging, the segmented endoscope may appear to extend beyond the bronchial wall boundaries. Despite this, the autonomous agent successfully navigated to the same target destination as the human operators, demonstrating the efficacy of the proposed system.

Finally, we assessed procedural efficiency through total navigation time and the number of executed actions (Fig. \ref{fig:invivo}G). The autonomous system required a mean duration of $417.1 \pm 74.9$ s, which was significantly longer than the senior bronchoscopist ($176.7 \pm 52.5$ s, P = 0.016) but comparable to the junior doctor ($334.3 \pm 34.7$ s, P = 0.109). This extended duration is a physical safety constraint rather than a computational limit. Although inference requires only 6 ms, we enforce a 3-s execution window per step (Table \ref{tab:timing_analysis}). This programmatic delay ensures that the robotic kinematics remain sufficiently slow to guarantee tissue safety and prevent trauma during dynamic in vivo navigation. Despite this artificially constrained traversal speed, the autonomous agent executed an average of $240.6 \pm 28.2$ discrete actions, a figure statistically indistinguishable from both the senior ($228.6 \pm 24.5$, P = 0.359) and junior ($233.3 \pm 31.9$, P = 0.797) operators. This equivalence in action count demonstrates that the autonomous system generates trajectories with a directness and economy of motion on par with clinical experts, effectively avoiding redundant maneuvers throughout the procedure.

To decouple decision-making from physiological constraints, we analyzed the frequency distribution of control primitives by projecting the action space onto two axes: lateral adjustment rate (steering frequency) and forward progression rate (longitudinal speed) (Fig. \ref{fig:invivo}H). Distinct behavioral clusters emerged from this analysis. The senior surgeon exhibited the highest magnitude in both metrics, reflecting superior proficiency in both rapid advancement and directional control.
Notably, a divergence was observed between the junior surgeon and the autonomous agent. While the junior surgeon surpassed the agent in forward progression rate, the agent outperformed the junior surgeon in lateral adjustment rate. We attribute this discrepancy to fundamental differences in actuation and processing. For human operators, forward progression is often a cognitively low-demand task allowing for continuous actuation; however, the autonomous system operates under a uniform computational cadence, processing forward and lateral states with equivalent inference latency, which creates a ceiling on its longitudinal speed. 
Conversely, in the decision-critical lateral adjustments, the junior surgeon exhibits intermittent hesitation, whereas the autonomous agent sustains consistent algorithmic throughput. As a result, the robot responds more efficiently during complex steering tasks, despite operating at a lower linear advancement speed overall.

\section{Discussion}

Accurate navigation remains a major barrier in robotic bronchoscopy, particularly in distal airways where anatomy is narrow, deformable, and visually repetitive. Existing robotic systems often depend on external localization hardware, such as electromagnetic (EM) tracking or shape sensing, that can increase procedural complexity and remain vulnerable to mismatch between preoperative images and intraoperative anatomy. 
In this study, we show that autonomous bronchoscopic navigation can be achieved using endoscopic vision and preoperative CT, without relying on external tracking sensors during navigation. 
Across a high-fidelity phantom, ex vivo porcine lungs, and a live porcine model under active respiration, the proposed framework achieved consistent distal target access with performance approaching that of a senior expert.

A central challenge in bronchoscopic autonomy is that the bronchial tree is difficult to localize geometrically in a robust way. Airway branches are visually similar, the endoscopic field of view is limited, and deformation, mucus, and motion can degrade the image stream. 
Rather than treating navigation primarily as an explicit pose-estimation problem, our framework reformulates it as sequential visual alignment between the live endoscopic view and a precomputed series of virtual targets. 
This design shifts the control problem from recovering a globally precise geometric state to maintaining local alignment with a planned visual trajectory. 
By converting a long-horizon navigation task into a sequence of local visual decisions, the framework maintained progress along planned trajectories while remaining responsive to the intraoperative scene.

The hierarchical agents were important for making this formulation practical in long-horizon navigation. The short-term reactive agent handled continuous low-latency maneuvers, enabling real-time adjustment to local scene changes.  The Long-term Strategic Agent was invoked only at anatomically ambiguous decision points or under predefined failure conditions, where local visual evidence alone was more likely to be ambiguous. 
This separation of timescales allowed the system to combine rapid local control with higher-level decision support when local image evidence was insufficient.

The world model provided an additional layer of decision support by evaluating candidate actions before execution. Instead of relying exclusively on a feedforward policy output, the system could simulate short visual rollouts and compare predicted future states with the intended virtual target. 
This mechanism was most valuable when the reactive and strategic modules produced conflicting recommendations, especially in visually similar branch regions where a locally plausible action could still lead the robot away from the planned route. 
In this sense, the world model functioned as a predictive critic that improved consistency under challenging regions such as branch points and under moderate visual perturbation.

The translational implication of these results is that autonomous navigation may be achievable in standard bronchoscopic workflows: preoperative CT for planning and endoscopic imaging for intraoperative navigation. This design may reduce dependence on specialized tracking hardware while preserving the ability to reach peripheral targets in airways. 
In the in vivo porcine experiments, the autonomous system achieved CBCT-verified endpoint accuracy and terminal-view similarity that were comparable to the inter-operator variation between a senior and a junior bronchoscopist. Although this does not establish clinical equivalence, it supports the feasibility of sensor-free intraoperative navigation in a realistic preclinical model.

Several limitations should be considered. First, the current system remained slower than expert teleoperation. In our experiments, this was driven largely by a deliberately imposed fixed execution window after each command to limit tissue trauma and ensure stable physical actuation, rather than by the core inference latency of the reactive policy itself. Even so, reducing end-to-end execution time will be important for practical use. 
Second, the system remained vulnerable to severe visual failure modes. Partial degradation from mucus or lens contamination could often be tolerated, but persistent lens fouling or complete occlusion of the target lumen still caused failures, highlighting an inherent weakness of a vision-only strategy under extreme visual corruption. 
Finally, our current validation focuses on reaching the target; the subsequent tasks of biopsy sampling and tool-tissue interaction require a different set of dexterous skills that our current navigation policy does not encompass.

In conclusion, these results show that pure-vision control, combined with hierarchical agents and world model, can support robust autonomous bronchoscopic navigation in deformable airways. By reframing bronchoscopy as a sequential visual-alignment problem, the proposed system reduces reliance on external localization hardware while maintaining strong performance across increasingly realistic experimental settings. This provides a foundation for future endoluminal robotic systems that are more adaptive and better suited to the dynamic conditions of the living airway.


\clearpage 

%
\bibliography{science_template} 
\bibliographystyle{sciencemag}


\section*{Acknowledgments}

\paragraph*{Competing interests:}
 The authors declare there are no competing interests.
\paragraph*{Data and materials availability:}
Data and code will be made publicly available after publication.


\subsection*{Supplementary materials}
Materials and Methods\\
Supplementary Text\\
Figs. S1 to S17\\
Tables S1 to S2\\
Movie S1 to S5\\


\newpage


\renewcommand{\thefigure}{S\arabic{figure}}
\renewcommand{\thetable}{S\arabic{table}}
\renewcommand{\theequation}{S\arabic{equation}}
\renewcommand{\thepage}{S\arabic{page}}
\setcounter{figure}{0}
\setcounter{table}{0}
\setcounter{equation}{0}
\setcounter{page}{1} 


\begin{center}
\section*{Supplementary Materials for\\ \scititle}


	Junyang Wu,
	Mingyi Luo,
    Fangfang Xie,
    Minghui Zhang,
    Hanxiao Zhang,\\
    Chunxi Zhang,
    Junhao Wang,
    Jiayuan Sun$^{\ast}$,
    Yun Gu$^{\ast}$,
    Guang-Zhong Yang$^{\ast}$\\ 
\small$^\ast$Corresponding author. Email: example@mail.com\\
\end{center}

\subsubsection*{This PDF file includes:}
Materials and Methods\\
Supplementary Text\\
Figures S1 to S17\\
Tables S1 to S2\\
Captions for Movies S1 to S5\\

\subsubsection*{Other Supplementary Materials for this manuscript:}
Movies S1 to S5\\

\newpage


\subsection*{Materials and Methods}

\subsubsection*{Study Design and Experimental Platform}

All experiments were conducted using a custom-developed robotic bronchoscopy platform. The system features a flexible catheter with a 4.2-mm outer diameter, a working channel, and a distal CMOS camera ($392 \times 392$ resolution). 
The robot accepted seven discrete commands: forward, backward, bend left, bend right, bend up, bend down, and subgoal switch. Each translational command corresponded to a 2-mm step, and each bending command corresponded to a 9$^\circ$ articulation step. The default actuator settings used across all experiments are summarized in Table~\ref{tab:core_robot_config}. At test time, the controller receives endoscopic observations and the virtual target derived from the preoperative CT plannning path, and outputs one of seven discrete actions.

Computation was performed on a workstation equipped with dual NVIDIA RTX 4090 GPUs. The Short-term Reactive Agent ran continuously on one GPU, whereas the World Model was executed on the second GPU only when invoked for conflict resolution. The control loop was synchronized to physical actuation rather than to network latency. In each cycle, the system captured an endoscopic frame ($\sim 30$ ms), computed one action ($\sim 6$ ms), and then enforced a fixed 3.0-s action executing period before the next cycle. This dwell time was introduced as a safety constraint to ensure mechanical settling before a subsequent command.

\subsubsection*{Data Acquisition and Labelling}

The training corpus combined data from three domains: in vitro airway phantoms, ex vivo porcine lungs, and in vivo porcine lungs. In this study, an ``instance'' denotes one labeled sample consisting of an endoscopic observation, the virtual target, and one expert-annotated valid action for that state. The full corpus comprised 12,218 phantom instances, 5,414 ex vivo porcine instances, and 16,206 in vivo porcine instances. The phantom set was composed of 2,865 manually annotated expert-demonstration instances, 8,275 synthetic style-transferred instances, and 1,078 manually collected real-goal instances.
To scale up the data, we used two complementary data-acquisition paradigms: task-directed expert demonstrations and style-transfer-augmented generation. Samples from the same continuous navigation sequence were kept within the same partition during retrospective development to reduce leakage between model training and held-out evaluation.

\paragraph{Task-Directed Expert Demonstrations} In the primary annotation mode, preoperative segmentation and path-planning algorithms were employed to render virtual images, which served as navigational targets. Expert operators manually teleoperated the robotic system, using these virtual images as visual goals. The expert’s applied kinematic commands (actions) were recorded synchronously with the visual observations to establish ground-truth labels. However, we identified that this classical imitation learning paradigm inherently lacks behavioral diversity; the deterministic nature of the virtual targets and the highly consistent navigational habits of expert operators resulted in a narrow state-action distribution, strictly limiting the scalability of the dataset.

\begin{figure} 
	\centering
	\includegraphics[width=\textwidth]{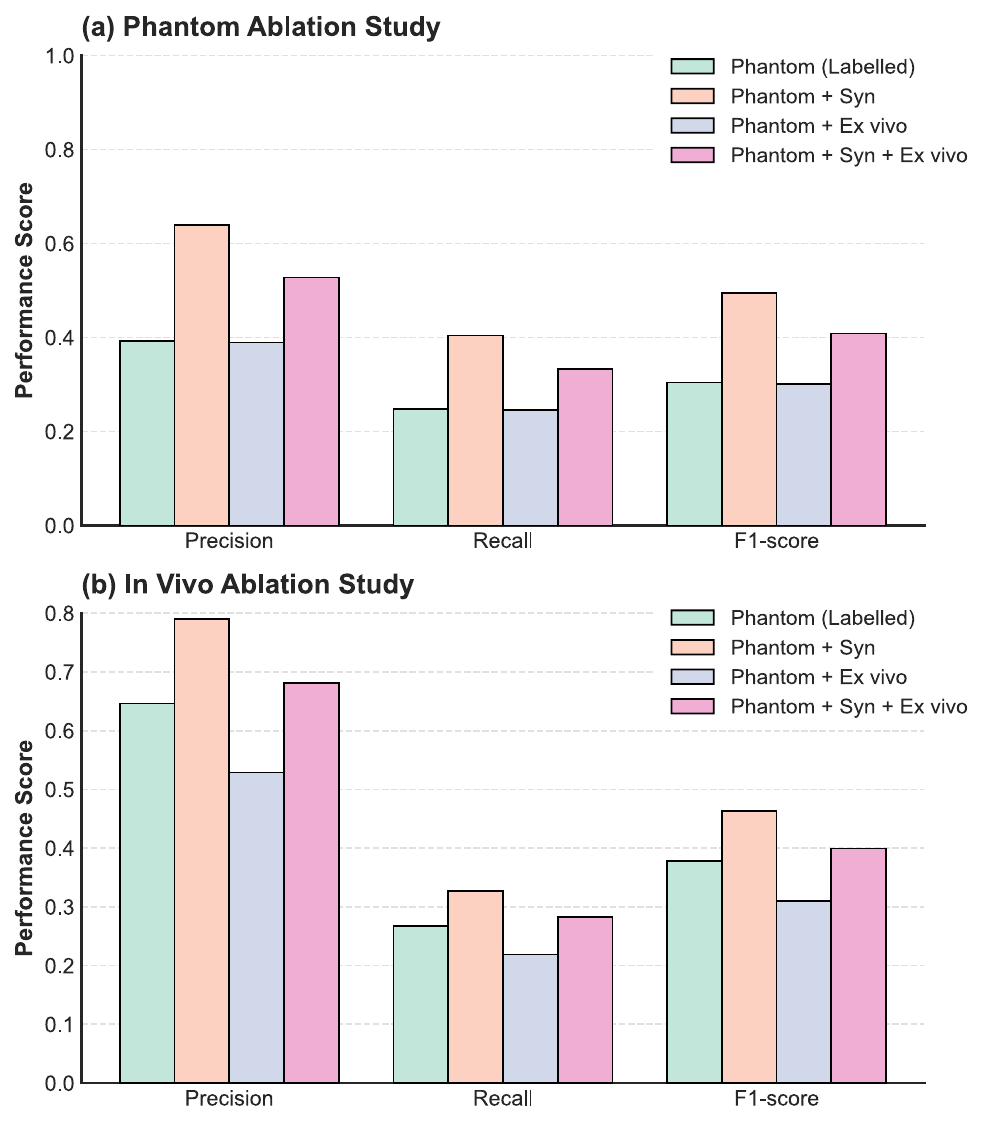} 

	\caption{\textbf{Ablation study on in vivo phantom and retrospective in vivo data.} We evaluate model performance on a gold-standard retrospective in phantom and in vivo dataset using precision, recall, and F1-score under increasing training data scales. When trained solely on expert-annotated phantom data, the model exhibits poor generalization. Incorporating CycleGAN-based synthetic data leads to a substantial performance improvement, highlighting the importance of data scaling via style transfer. Further inclusion of ex vivo porcine lung data yields a modest additional gain.}
	\label{fig:supp14} 
\end{figure}

\paragraph{Style-Transfer-Augmented Generation} To overcome this limitation and enrich the state-action space, we proposed a data generation strategy based on style transfer. Experts were instructed to perform unconstrained, randomized exploratory maneuvers within the phantom environment—executing continuous multi-axis articulations and intentional wall contacts. We formulated temporal data triplets consisting of a pre-action observation image, a post-action target image, and the intermediate executed action.
Since the target images during the inference phase belong to the virtual domain, we applied a CycleGAN \cite{zhu2017unpaired} to perform domain translation, mapping the real post-action images into the virtual target domain. This synthesis pipeline successfully scaled the phantom dataset from the initial 2,865 manually annotated cases to a robust suite of 12,218 instances (2,865 manual, 8,275 synthetic, and 1,078 real-goal manual). As demonstrated in our ablation studies (Fig. \ref{fig:supp14}), integrating this CycleGAN-based synthetic data significantly mitigates the reality gap, fundamentally enhancing the robustness and cross-domain generalization capabilities of the underlying model.

\paragraph{Capturing and labelling the ex vivo and in vivo datasets} 
Both the ex vivo and in vivo datasets were collected under the task-directed expert demonstration paradigm, because unconstrained randomized exploration was not appropriate in biologic specimens. For the ex vivo dataset, an expert manually teleoperated the robotic bronchoscope in isolated porcine lungs and navigated along preplanned routes toward the designated targets. The endoscopic observations, virtual targets, and executed actions were recorded as labeled training samples. 
For the in vivo dataset, we used retrospective real endoscopic images acquired during expert-operated procedures. For each case, we reprocessed the corresponding preoperative CT to reconstruct the airway model and render the associated virtual target sequence, after which an expert reviewed each real-virtual image pair and assigned the most appropriate action label.

\subsubsection*{Short-term Reactive Agent Implementation}

The core learned controller was the Short-term Reactive Agent, which produced one discrete navigation action per control cycle. At time step $i$, the controller received a short visual history $o_{i-P:i} = \{o_{i-P}, \ldots, o_i\}$ from the bronchoscope camera together with the active virtual subgoal $v_j$. The subgoal $v_j$ was the pre-rendered virtual bronchoscopic view associated with the current waypoint along the planned centerline. For inference, the current observation and goal image were resized to $85 \times 64 \times 3$ before being passed to the vision backbone. The policy network $\pi_{\theta}$ then predicted the next action,conditioned on the observation history and active goal:
\begin{equation}
    \hat{a}_{i:i+K} = \pi_{\theta}(o_{i-P:i}, v_{j})
\end{equation}
where $P=1$ denotes the observation-history length and $K=1$ denotes one-step action prediction in the present implementation.

For each RGB frame $o_i \in \mathbb{R}^{H \times W \times 3}$, an EfficientNet-B0 backbone~\cite{koonce2021efficientnet} extracted a visual representation, followed by spatial pooling to produce a 512-dimensional embedding. The observation and goal embeddings were concatenated and processed by a decoder-only Transformer~\cite{han2021transformer}. Following ViNT~\cite{shah2023vint}, the Transformer used $L=4$ self-attention blocks, $n_h=4$ attention heads, and hidden width $d_{model}=2048$. A final fully connected layer projected the decoded representation to the seven-action command space: forward, backward, bend left, bend right, bend up, bend down, and subgoal switch. ``Subgoal switch'' indicated that the current waypoint had been reached and that the controller should advance to the next virtual target along the preoperative route. The reactive policy was trained by supervised imitation learning using cross-entropy loss against expert-labeled actions.

\subsubsection*{Long-term Strategic Agent Implementation}
The Long-term Strategic Agent serves as a high-level supervisor, integrating geometric priors with semantic reasoning to resolve navigational ambiguities. This agent comprises two distinct modules: a Foundation Model-based planner and a geometric preoperative guidance module derived from the planned virtual trajectory.

\begin{figure} 
	\centering
	\includegraphics[width=\textwidth]{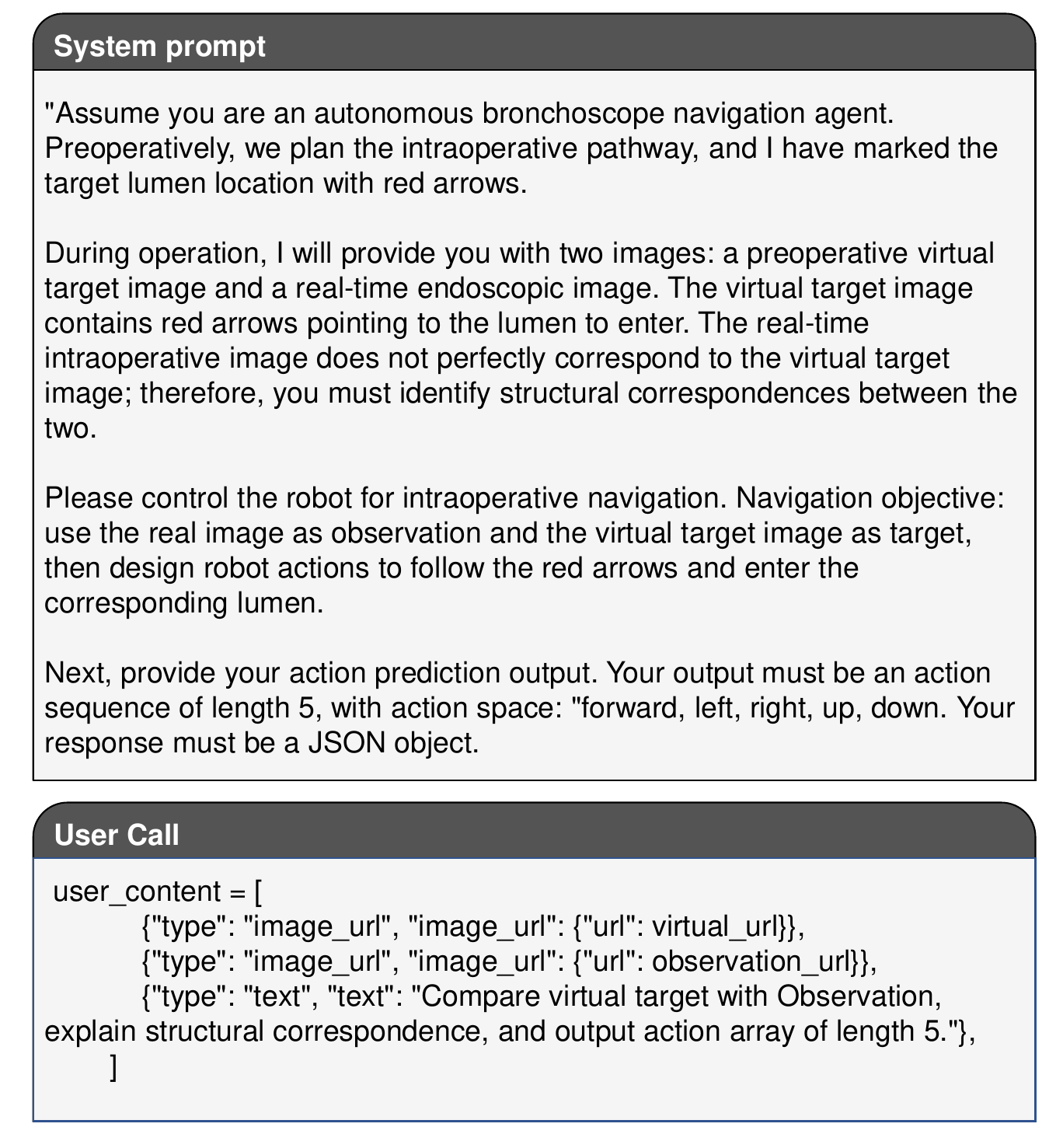} 

	\caption{\textbf{Prompts for the LLM-guided agent.} The prompt design comprises a fixed system prompt and a dynamically generated user prompt instantiated at each inference step.}
	\label{fig:supp10} 
\end{figure}

\begin{figure} 
	\centering
	\includegraphics[width=\textwidth]{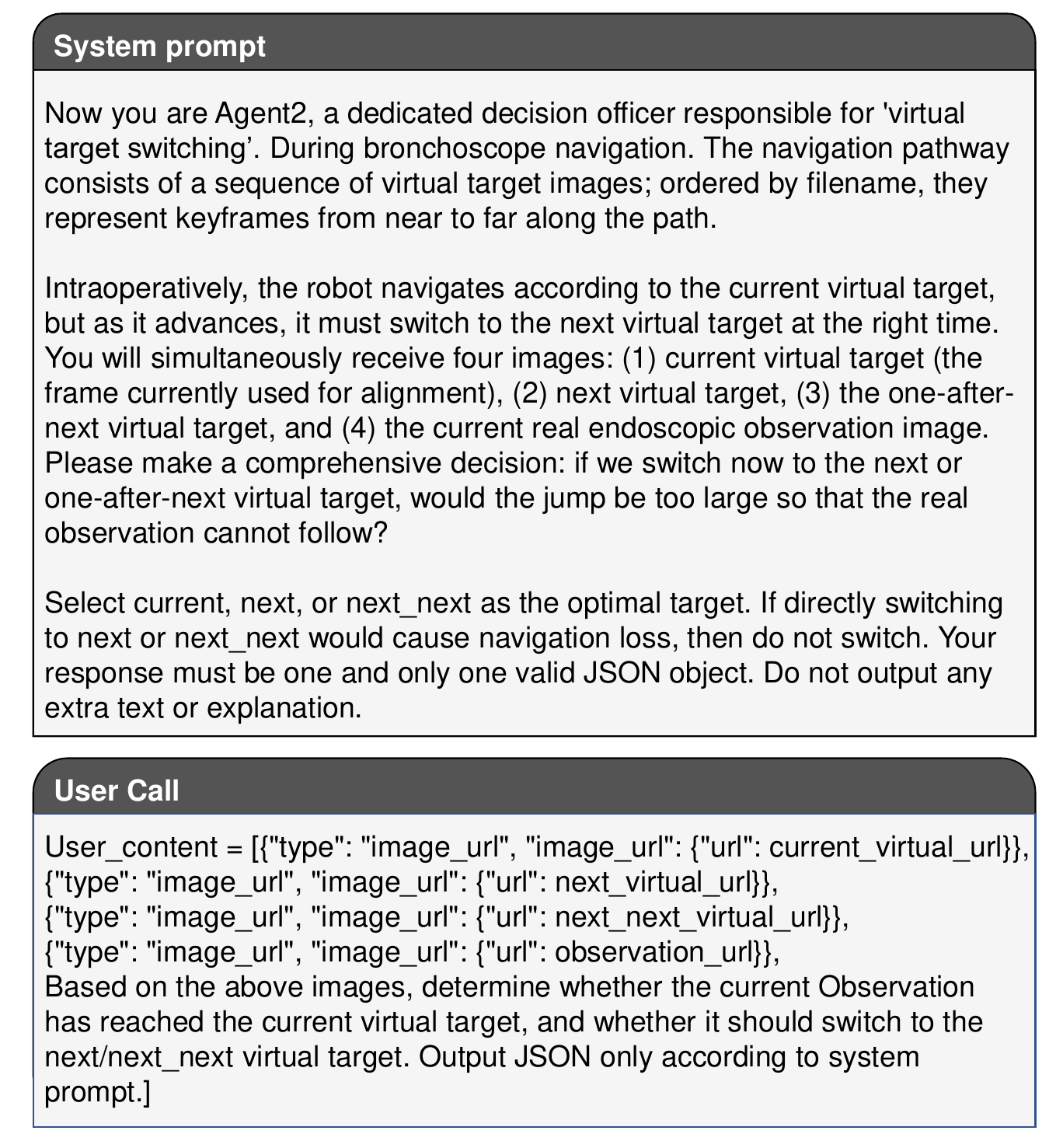} 

	\caption{\textbf{Prompts for the target switch agent.} The prompt design comprises a fixed system prompt and a dynamically generated user prompt instantiated at each inference step.}
	\label{fig:supp11} 
\end{figure}

\textbf{LLM Guidance:} We utilized a Large Multimodal Model (Gemini-3-flash, Google DeepMind) to provide semantic guidance specifically at high-ambiguity branch points. 
The module received three inputs: the current intraoperative observation, the corresponding virtual target, and engineered prompts. 
Based on these inputs, the model synthesizes a discrete, five-step action sequence, which serves as an action candidate to guide the robot's subsequent navigational maneuvers.
To ground the model's reasoning, we implemented a \textit{Dual-Prompting Strategy}. For the \textit{visual prompting}, virtual reference images were rendered with a visual indicator (a red directional arrow) aligned with the extracted centerline, explicitly marking the target lumen (Fig. \ref{fig:supp7}). For the \textit{textual prompting}, we supplied the model with a structured system prompt defining its role as a bronchoscopic navigation agent. The model was instructed to semantically align the intraoperative view with the arrow-guided virtual target to identify structural correspondences. The output was constrained to a JSON format containing a specific navigational command (Forward, Backward, Left, Right, Up, Down) and a reasoning justification. The detailed textual prompting are shown in Fig. \ref{fig:supp10} and \ref{fig:supp11}.
To balance computational cost with navigational precision, this module is selectively triggered at the penultimate bronchial bifurcation, where the branching topology becomes singular and locally indistinct. Due to the inference latency, the robotic system executes a temporary hold during query processing.

\textbf{Preoperative Guidance:} This module established a deterministic action prior from the planned virtual route. During preoperative planning, we generated a contiguous sequence of virtual target frames and their associated 6-DoF poses. For each adjacent frame pair $(v_i, v_{i+1})$, we computed the relative pose transformation and projected it into the discrete action space to obtain one candidate action. These candidate actions were logged offline along the planned route. 
Under the assumption that the virtual target and the real intraoperative observation remain spatially proximate during navigation, this sequence of actions derived strictly from the virtual environment can provide directional priors for the robot.

To operationalize this mechanism, we preoperatively log the discrete navigational actions connecting all adjacent virtual frame pairs. Intraoperatively, the control system maintains a rolling action buffer. Upon system query, the agent aggregates the action history from the preceding 10 frames, employing a temporal majority voting scheme to output the dominant action as the definitive guidance command. Notably, this module generates action candidates relying exclusively on the preoperative virtual trajectory, strictly decoupling from the real observation. 
This modality-isolated paradigm mitigates the inherent fragility of pure-vision control policies, providing a failsafe when real intraoperative views are severely corrupted by visual artifacts or anatomical occlusions.

The preoperative module is conditionally activated under two predefined failure specifications: (i) \textit{Stagnation Recovery:} If the primary policy predicts continuous rotational actions (bending) for 8 consecutive frames without concurrent translation, the system calls the preoperative guidance and enforces a forward action after that. (ii) \textit{Forward Bias Correction:} When three consecutive forward actions are detected, the system invokes the preoperative guidance module to trigger the necessary turning maneuvers.

\subsubsection*{World model Implementation}

We built a predictive latent video world model to arbitrate action conflicts between the short-term reactive agent and long-term strategic agent. Let $o_{i-h:i}$ denote the previous $h+1$ RGB frames and $a_{i:i+K-1}$ a candidate $K$-step action sequence. A variational autoencoder (VAE)~\cite{kingma2013auto} encoded each frame into a latent tensor $z_i = \mathrm{Enc}(o_i)$. Conditioned on the history latents $z_{i-h:i}$ and candidate actions $a_{i:i+K-1}$, the model predicted future latent features $\hat{z}_{i+1:i+K}$, which were decoded by a frozen VAE decoder to generate predicted future observations $\hat{o}_{i+1:i+K} = \mathrm{Dec}(\hat{z}_{i+1:i+K})$.

During training, history tokens were treated as conditioning inputs and only future prediction tokens were perturbed. For predictive latent features $\hat{z}^{(0)}$, we sampled a diffusion step $t \in [1,T]$ and obtained the perturbed latent feature $\hat{z}^{(t)} = \sqrt{\bar{\alpha}^{(t)}}\hat{z}^{(0)} + \sqrt{1-\bar{\alpha}^{(t)}}\epsilon^{(t)}$, where $\epsilon^{(t)} \in \mathcal{N}(0,\mathbf{I})$.

For the network architecture, we follow IRASim~\cite{zhu2025irasim}, which applies Diffusion Transformers (DiT)~\cite{peebles2023scalable} containing spatial attention blocks and temporal attention blocks, and utilize adaptive layer normalization for conditioning. 
After $L$ DiT blocks, the final linear layer outputs the predicted noise $\hat{\epsilon} =  \epsilon_{\theta}(\hat{z}^{(t)}, t, z_{i-h:i}, a_{i:i+K-1})$. The loss function for the predicted noise ~\cite{ho2020denoising} is 
\begin{equation}
    L_{diff} = \mathbb{E}_{t,\epsilon}[||\epsilon - \epsilon_{\theta}(\hat{z}^{(t)}, t, z_{i-h:i}, a_{i:i+K-1})||_{2}^{2}]
\end{equation}

During inference, for each candidate action, the model generated a short rollout from Gaussian noise $x^{(T)} \in \mathcal{N}(0,\mathbf{I})$ using iterative denoising:
\begin{equation}
    x^{(t-1)} = \frac{1}{\sqrt{\alpha^{(t)}}}(x^{(t)} - \frac{1-\alpha^{(t)}}{\sqrt{1-\bar{\alpha}^{(t)}}}\epsilon_{\theta}(x^{(t)},t, a))
\end{equation}
where $a$ is the candidate action.

The predicted features were represented as $\hat{z}_{i:i+K}=x^{(0)}$ and decoded into future image sequences. Candidate actions were ranked by LPIPS distance between the predicted future view and the active virtual target, and the action with the lowest perceptual distance was selected.

\subsubsection*{Pre-operative image processing steps}

\begin{figure} 
	\centering
	\includegraphics[width=\textwidth]{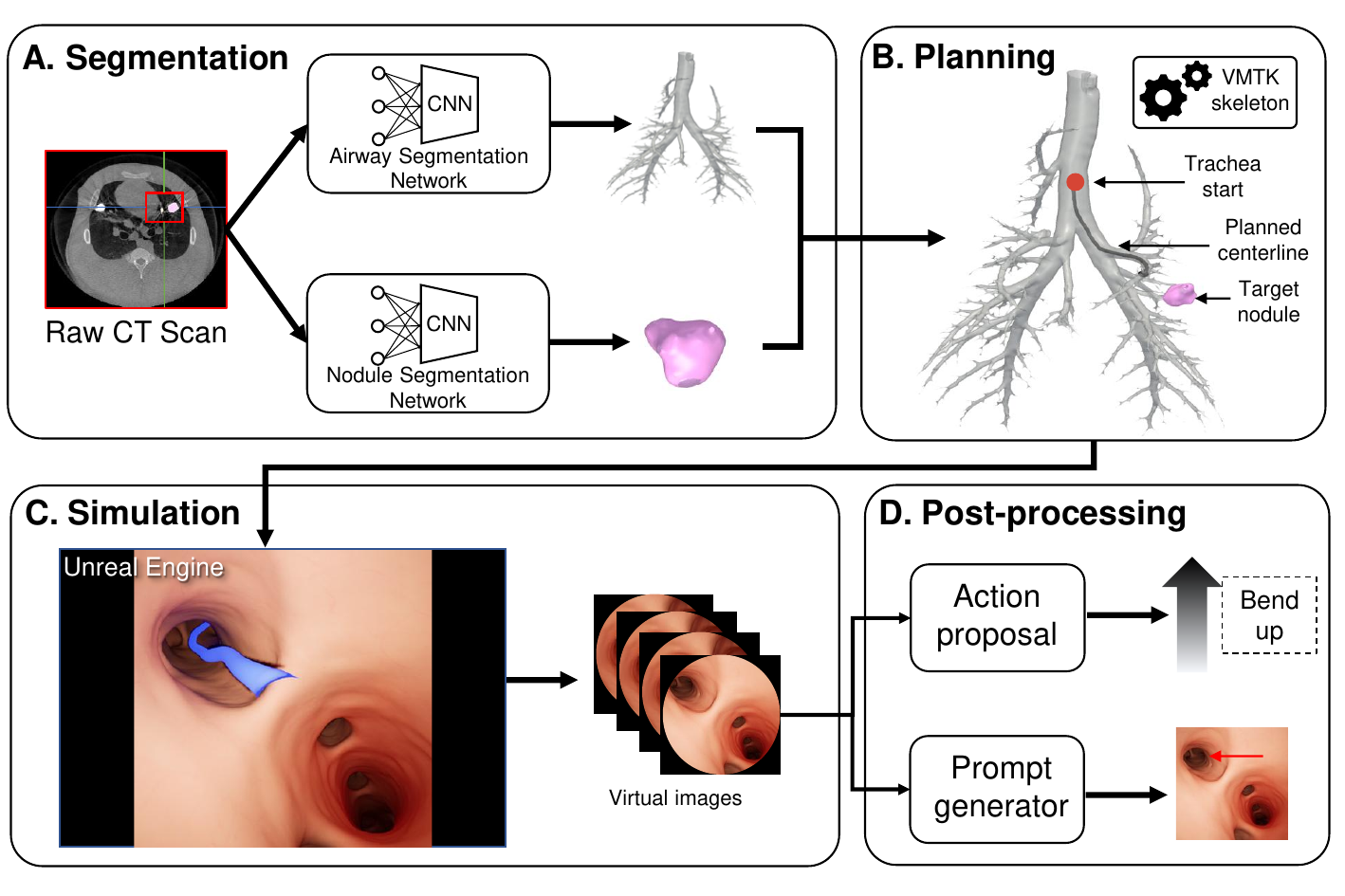} 

	\caption{\textbf{Virtual image generation pipeline.} (A) CT segmentation pipeline: raw CT volumes are processed by dedicated airway and nodule segmentation networks to reconstruct 3D meshes of the bronchial tree and target nodules. (B) Path planning via VMTK skeletonization: using the trachea as the start point and an intraluminal point near the target nodule as the endpoint, a navigation centerline is generated. (C) Virtual rendering: the segmented airway mesh and planned centerline are imported into a virtual engine to synthesize virtual images serving as virtual targets. (D) Post-processing: action candidates and prompts are generated. Action candidates are obtained by computing the relative pose between adjacent virtual images and mapping it to one of six discrete actions. Prompts indicate the lumen aligned with the centerline vanishing direction, with the target lumen highlighted by a red arrow.}
	\label{fig:supp7} 
\end{figure}

As shown in Fig. \ref{fig:supp7}, pre-operative planning was performed from the patient CT volume through a fully automated image-processing pipeline. First, the raw CT data were processed by two dedicated segmentation networks to reconstruct the bronchial tree and the target nodule, respectively, thereby generating patient-specific 3D meshes of the airway lumen and lesion. These segmentations established the anatomical substrate used for downstream path planning and virtual target generation.

Next, a navigation centerline was computed on the segmented airway tree using Vascular Modeling Toolkit (VMTK)-based skeletonization. The trachea was specified as the start node, and an intraluminal point nearest the target lesion was specified as the terminal node, yielding a continuous centerline. The airway mesh and planned centerline were then imported into an Unreal Engine-based virtual rendering environment to synthesize a sequence of virtual bronchoscopic views along the path. Each rendered frame served as an image-space subgoal and was associated with a corresponding pose on the pre-planned trajectory.

To support downstream guidance, the rendered sequence was further post-processed to generate both action priors and visual prompts. Specifically, the relative pose between adjacent virtual views was mapped to one of six discrete control actions, producing a candidate action set for each subgoal transition. In parallel, the lumen aligned with the centerline vanishing direction was highlighted in the virtual target image to provide an explicit directional cue. During autonomous navigation, the system used only the live monocular endoscopic observation and these pre-computed virtual subgoals, without external localization or tracking sensors.

\subsubsection*{Evaluation Metrics}

We predefined four primary navigation metrics: success rate, visual alignment consistency, CBCT endpoint distance, and action count.

\paragraph{Success rate.}
For each trajectory, success was determined under continuous supervision by an expert bronchoscopist. A trial was counted as successful only if the autonomous system (i) followed the same anatomical route as the expert-defined trajectory without entering an incorrect branch, and (ii) reached the same target bronchial structure as the expert endpoint. Any wrong-branch entry or failure to reach the matched target structure was counted as failure.

\paragraph{visual alignment consistency.}
To assess visual alignment consistency between expert and autonomous algorithm in the sensorless vision-only setup, we utilized the visual similarity between the robot's and the expert's final views as a proxy for determining orientation consistency. 
We employed the Structure Similarity Index Measure(SSIM) as a quantitative metric to evaluate the visual fidelity between the autonomous and expert views. 
Specifically, we recorded the full navigation trajectories for both the autonomous agent and the expert. To mitigate the impact of temporal mismatch at the destination, our analysis focused on the terminal phase, defined as the final 30 frames of each sequence. Within this window, we identified the specific pair of frames (one from the autonomous run and one from the expert) that yielded the highest structural similarity. The SSIM of this optimal pair was adopted as the metric for visual alignment consistency for that trajectory.

\paragraph{CBCT endpoint distance.}
Spatial endpoint agreement in the \textit{in vivo} study was measured in intraoperative CBCT space. The terminal bronchoscope tip positions of the autonomous system and the senior expert were identified in the same CBCT coordinate system, and endpoint deviation was computed as Euclidean distance:
\begin{equation}
d = \left\| \mathbf{p}_{\text{auto}} - \mathbf{p}_{\text{expert}} \right\|_2 .
\end{equation}

\paragraph{Action count.}
Action count was defined as the total number of discrete control commands executed during one trajectory. This count included all motion commands (forward, backward, bend left/right/up/down) and the subgoal-switch command.

\subsubsection*{Statistical Analysis}

Quantitative metrics are reported as mean $\pm$ standard deviation (SD) unless otherwise stated. Navigation success rate was defined as the percentage of trials in which the bronchoscope reached the target segment for that experiment. Procedure time was measured from the first executed control action to the terminal state. Action count was defined as the number of discrete commands generated by the agents, including target switch actions that did not actuate the hardware.
For all comparisons, trajectories were paired by anatomical target because the autonomous system and human operators were evaluated on the same planned routes. Accordingly, paired Student's $t$-tests were used for comparisons between the autonomous controller and the human operators. All reported $P$ values were two-sided, and $P < 0.05$ was considered statistically significant.


\subsection*{Supplementary Text}

\paragraph{Hardware Setup and Vision Acquisition} The vision feedback was acquired using a bronchoscope camera with a resolution of $392 \times 392$ pixels, operating at 30 frames per second (FPS). The camera intrinsic parameters were calibrated as follows: focal lengths $f_x \approx 290.0$, $f_y \approx 281.4$, and optical center $c_x \approx 203.3$, $c_y \approx 210.8$, with no specific distortion model assumed due to the specific lens design. The bronchoscope utilized in our experiments has an outer diameter of 4.2 mm and supports a maximum distal bending angle of 210°.
\paragraph{Action Space and Safety-Critical Control Strategy} We formulate the robotic control as a discrete position control problem, defining an action space that comprises seven distinct commands: forward, backward, bending left, bending right, bending up, bending down, and target switching. During physical execution, each translational command yields a precise 2-mm displacement, whereas each bending command actuates the distal servo by 9°. To guarantee intraoperative safety and strict visual-kinematic consistency, the system employs a synchronous, temporally gated control strategy. Although the neural network inference latency is negligible (~6 ms), we enforce a conservative, fixed 3-second temporal envelope for each mechanical execution phase. While the physical actuation may conclude well before this interval elapses, mandating the full 3-second window guarantees that the robot reaches a complete resting state before the subsequent control loop initiates. By strictly pausing new image acquisition and network forward passes during this envelope, the system inherently precludes dangerous command accumulation and ensures absolute kinematic stability.

\paragraph{In Vivo Validation Protocol}
All in vivo procedures were conducted in one conventional-grade white pig (40--45 kg). The study application was submitted on December 9, 2025, with acceptance number SHYS-No. 2025143, and received veterinary, laboratory, and IACUC approval (all approved on December 11, 2025; document code SHYS-SOP1-031-F003 A/2).

Seven target trajectories were prospectively defined for the in vivo study. Four trajectories terminated near implanted peripheral nodules, and three additional trajectories terminated at physician-selected distal bronchial targets without nodules. For each route, the autonomous system was initialized at the trachea and tasked with reaching the planned distal destination under the same controller configuration used in the benchtop studies.

\begin{figure} 
	\centering
	\includegraphics[width=\textwidth]{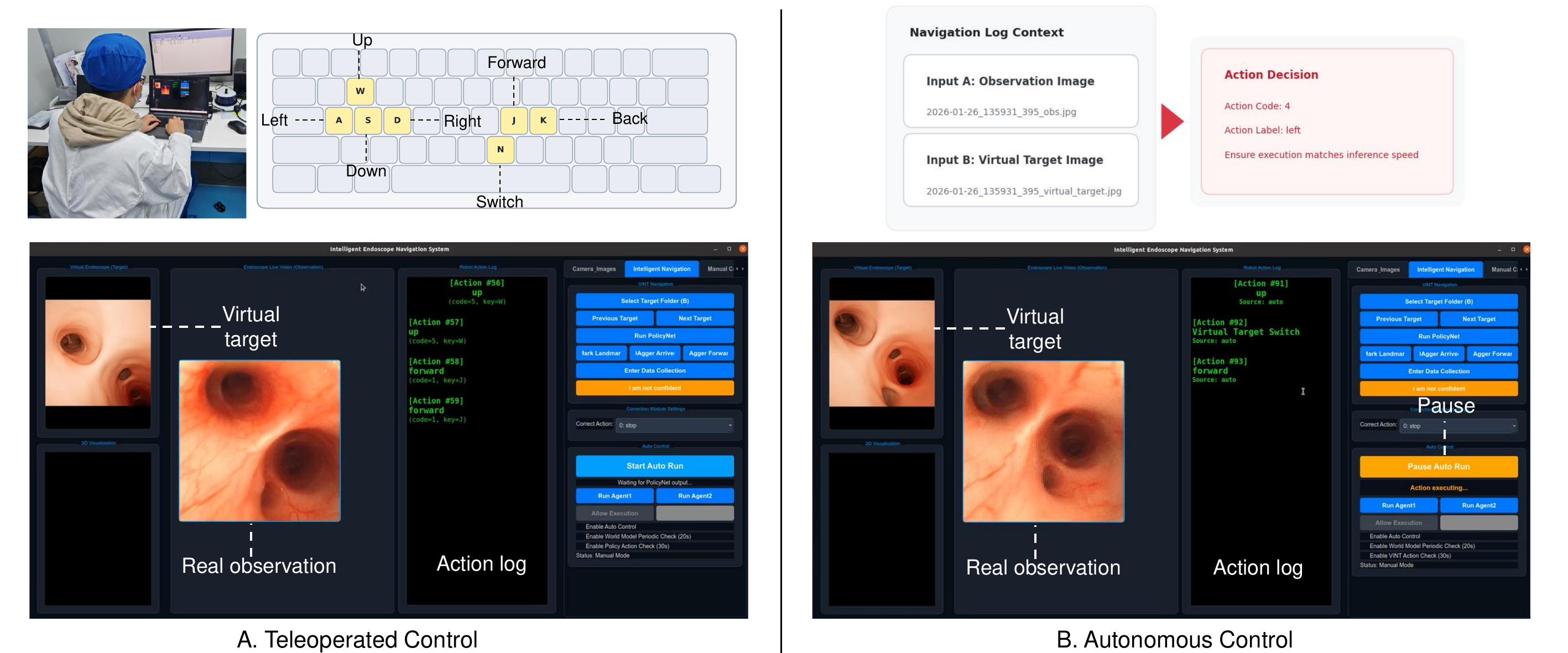} 

	\caption{\textbf{In vivo experimental settings for expert teleoperation and autonomous navigation.} (A) Teleoperated control. An expert remotely navigates the robot using a keyboard-based interface, with discrete commands for forward/backward motion, lateral and vertical adjustments, and target switching.
(B) Autonomous control. An autonomous agent executes navigation actions by jointly leveraging real-time observations and a virtual target representation, without human intervention.}
	\label{fig:supp8} 
\end{figure}

To provide clinical reference conditions, a senior bronchoscopist with more than 10 years of experience and a junior bronchoscopist each performed the same target-reaching tasks using the identical robotic hardware. Human operators controlled the platform through the same discrete keyboard command interface used by the autonomous system while viewing only the live endoscopic video feed (Fig. \ref{fig:supp8}). This design ensured that all conditions shared the same image source, robot, and action vocabulary.

\paragraph{Validation setting}

Although navigation success rate in physical experiments is the primary endpoint of this study, it is not practical as the sole criterion for iterative ablation analysis and model selection because full robotic navigation trials are comparatively time-consuming and resource-intensive. 
We therefore established an offline benchmark to provide a rapid and reproducible proxy for model quality before deployment in navigation experiments. Specifically, we quantified prediction consistency with human expert demonstrations, using this metric as a precursor to the ultimate evaluation of navigation success rate.

This evaluation protocol differed from the supervision used to construct the training dataset. During training data annotation, the expert provided only a single action label for each real-virtual pair, corresponding to the most appropriate immediate control decision at that state. During evaluation, however, multiple actions could be accepted for the same pair because the agent may still reach the same virtual target through different valid action sequences.

\begin{figure} 
	\centering
	\includegraphics[width=\textwidth]{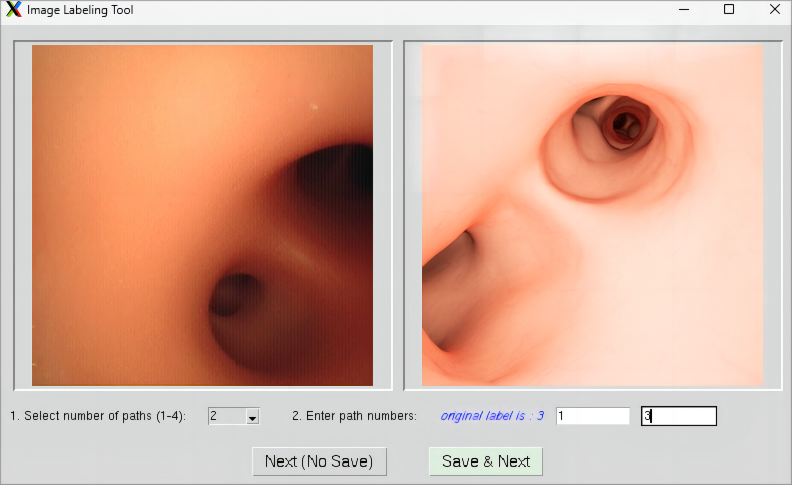} 

	\caption{\textbf{User interface for action annotation.} The interface presents real observations alongside corresponding virtual targets. Experts use the real image as the observation and the virtual target as the navigation goal, selecting one to four plausible actions as annotations, which are recorded as ground-truth labels.}
	\label{fig:supp12} 
\end{figure}

For this offline benchmark, the phantom evaluation set was collected from a separate phantom that was not used during training.
Within this phantom dataset, each real-virtual image pair was annotated with a subset of one to four ground-truth actions (Fig. \ref{fig:supp12}). This 1-to-4 cardinality arises because transitioning from a real observation to a virtual target may require a combination of up to four non-mutually exclusive action primitives: translation (forward/backward), horizontal bending (left/right), vertical bending (up/down), and target switching (`switch virtual target'). Diametrically opposed actions were excluded by construction. Consequently, a model prediction was considered correct if the predicted action fell within the annotated ground-truth subset for that state pair.



\begin{figure} 
	\centering
	\includegraphics[width=0.9\textwidth]{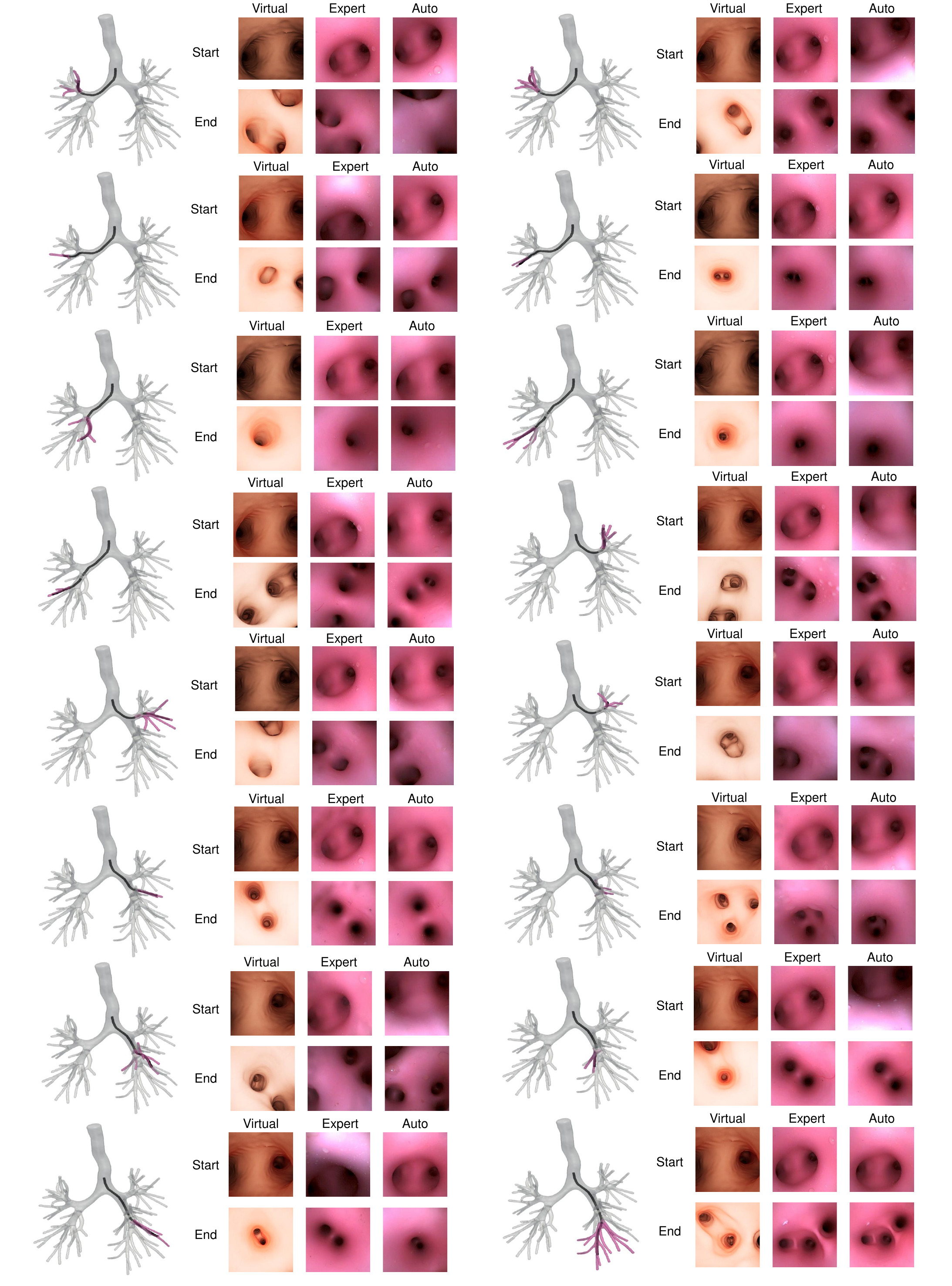} 

	\caption{\textbf{Representative trajectories from clean phantom experiments.} Sixteen representative cases are shown. For each case, the planned navigation trajectory within the bronchial phantom is overlaid, together with the endoscopic view at the onset of the experiment (start frame) and the corresponding image upon reaching the target location (end frame).}
	\label{fig:supp1} 
\end{figure}

\begin{figure} 
	\centering
	\includegraphics[width=\textwidth]{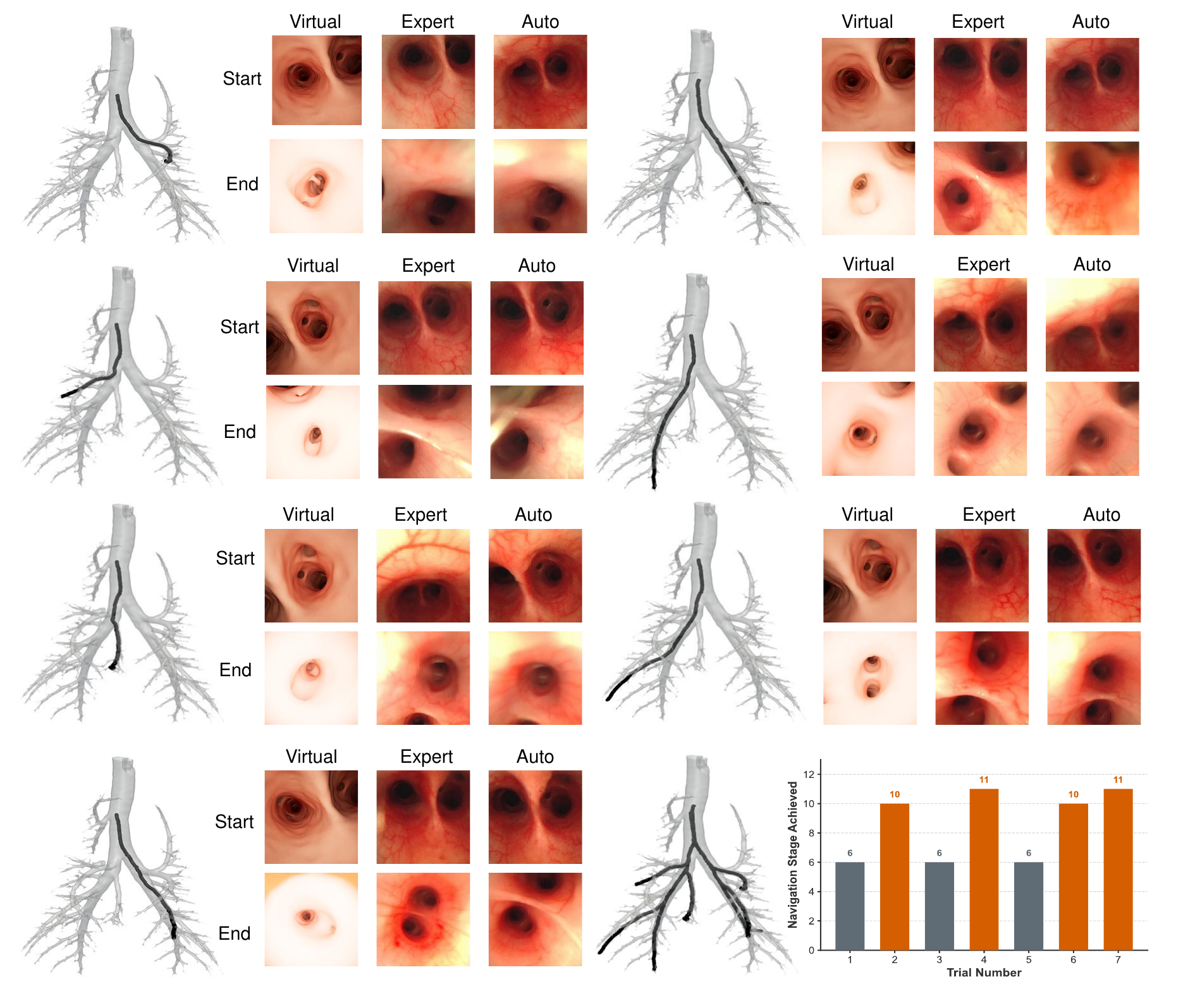} 

	\caption{\textbf{Representative trajectories from in vivo experiments.} Seven representative cases are shown. For each case, the planned navigation trajectory within the bronchial anatomy is overlaid, together with the endoscopic view at the onset of the experiment (start frame) and the corresponding image upon reaching the target location (end frame). In addition, a composite visualization summarizing all seven trajectories is provided, along with the terminal bronchial generation reached by each trajectory.}
	\label{fig:supp2} 
\end{figure}

\begin{figure} 
	\centering
	\includegraphics[width=\textwidth]{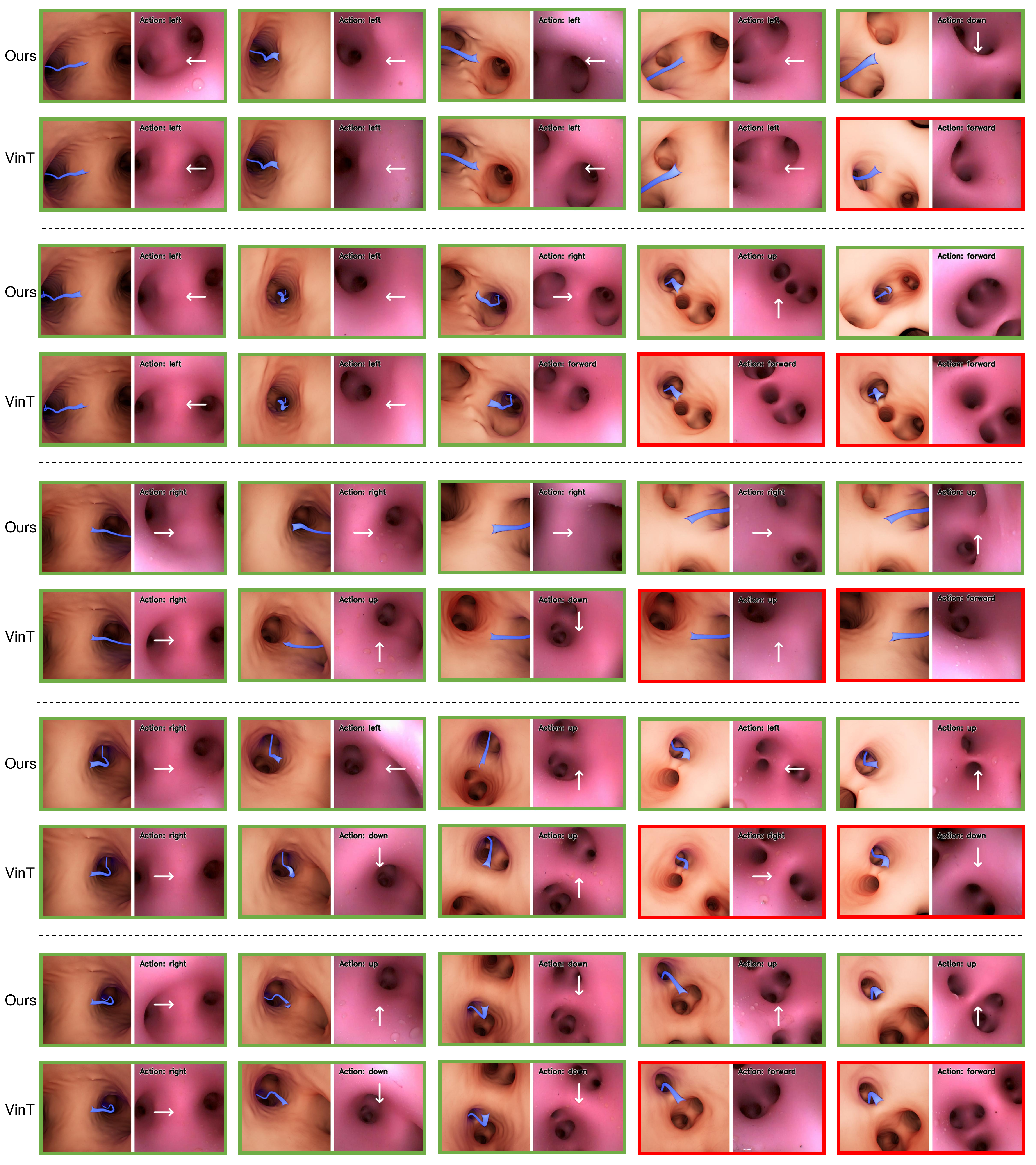} 

	\caption{\textbf{Image sequences along phantom navigation trajectories for the baseline and our method.} For each trajectory, five representative endoscopic images sampled from distinct bronchial branches are shown. Blue trajectories denote the centerline, directing toward the target lumen. Our method consistently follows the target navigation path and successfully reaches the intended branches. In contrast, the baseline (VinT) demonstrates reliable navigation in the early stages but frequently deviates from the target trajectory and becomes disoriented at deeper branches.}
	\label{fig:supp3} 
\end{figure}

\begin{figure} 
	\centering
	\includegraphics[width=\textwidth]{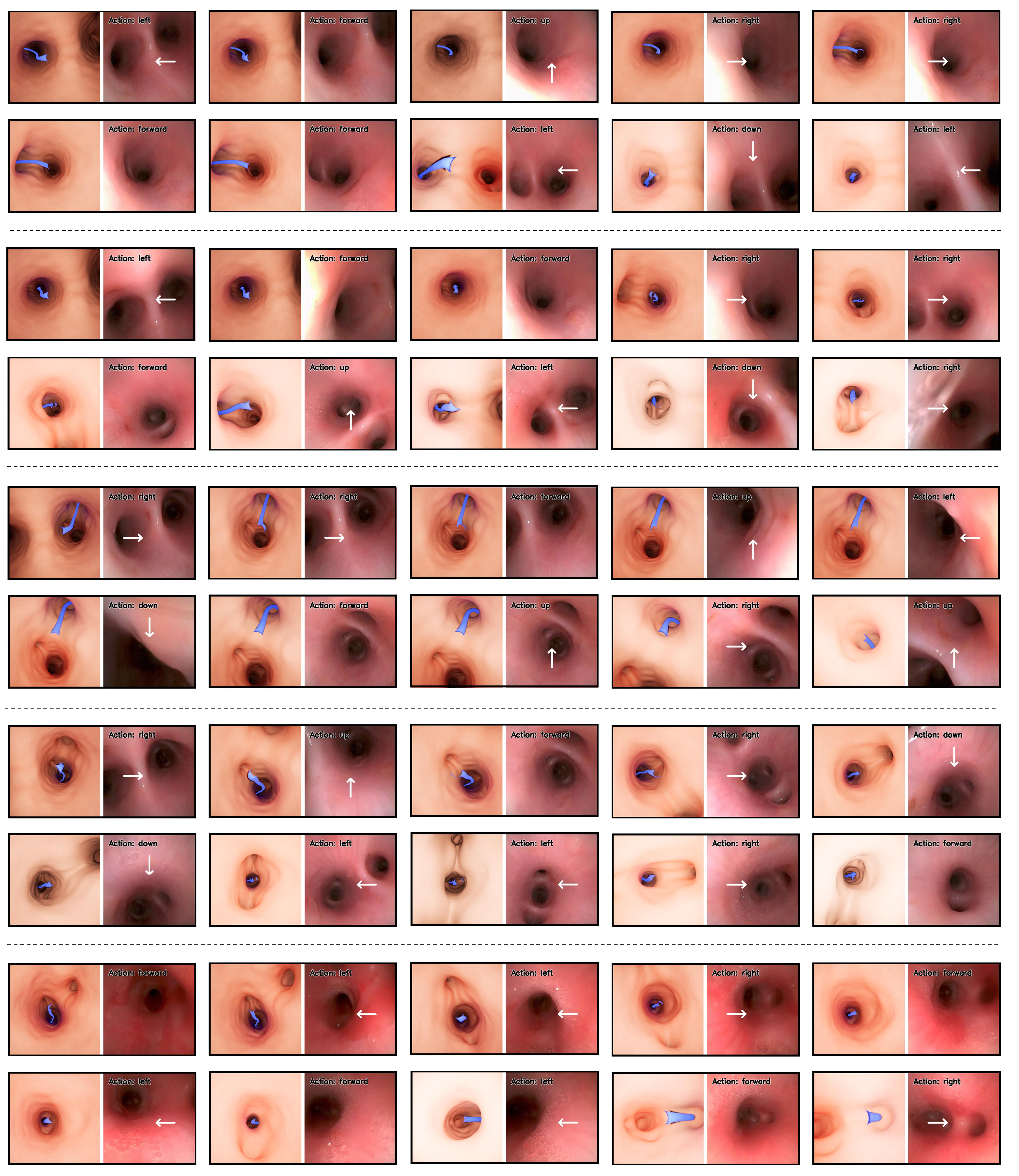} 

	\caption{\textbf{Image sequences along exvivo navigation trajectories for our method.} For each trajectory, ten representative endoscopic images sampled from distinct bronchial branches are shown. Blue trajectories denote the centerline, directing toward the target lumen. Our method consistently follows the target navigation path and successfully reaches the intended branches.}
	\label{fig:supp4} 
\end{figure}

\begin{figure} 
	\centering
	\includegraphics[width=\textwidth]{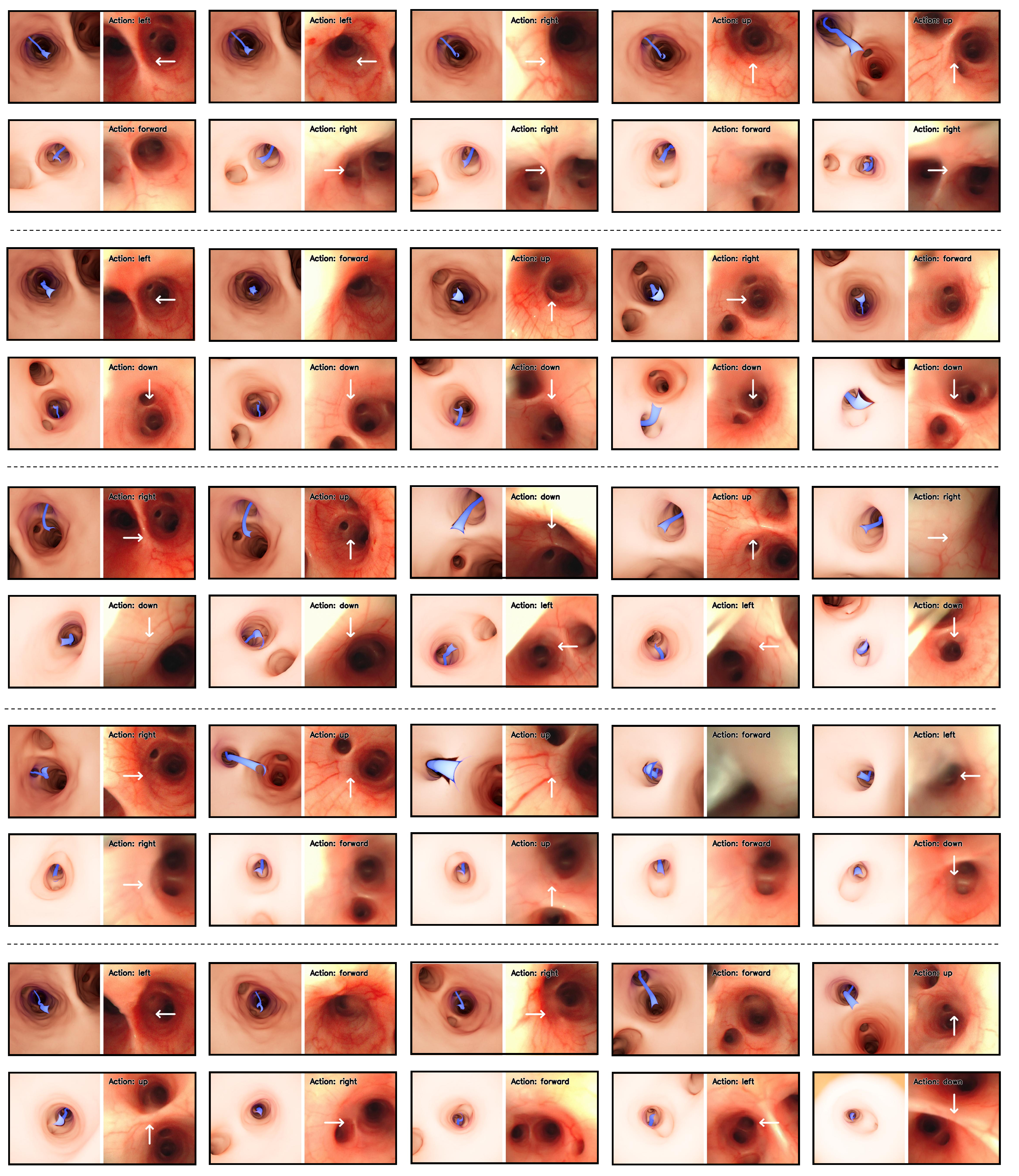} 

	\caption{\textbf{Image sequences along invivo navigation trajectories for our method.} For each trajectory, ten representative endoscopic images sampled from distinct bronchial branches are shown. Blue trajectories denote the centerline, directing toward the target lumen. Our method consistently follows the target navigation path and successfully reaches the intended branches.}
	\label{fig:supp5} 
\end{figure}

\begin{figure} 
	\centering
	\includegraphics[width=\textwidth]{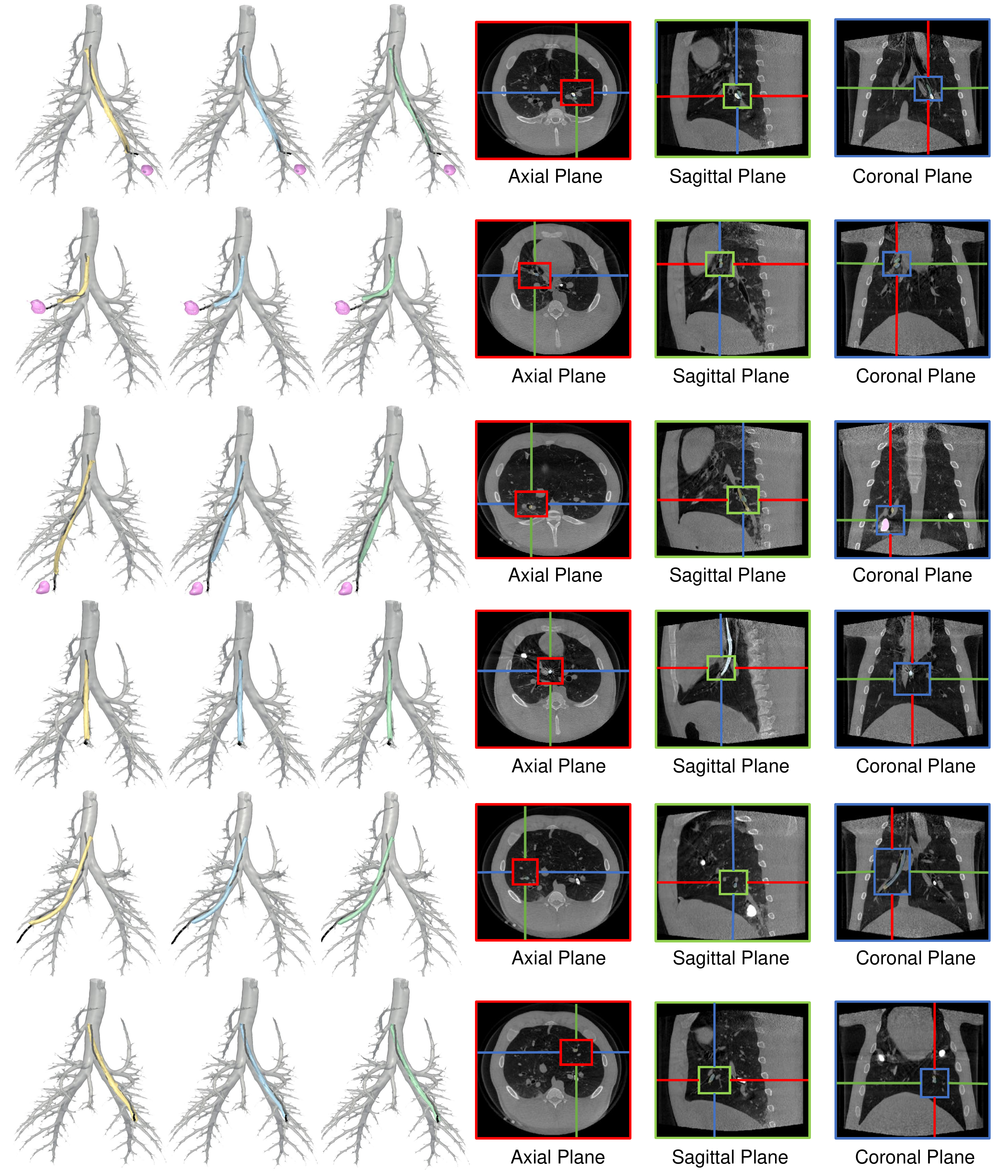} 

	\caption{\textbf{In vivo intraoperative CBCT visualization.} Left: CBCT-derived results from six in vivo experiments, including bronchial tree segmentation, centerline visualization, nodule localization, and the intraoperative bronchoscope position at the target site. Right: Raw intraoperative CBCT images, where three distinct colors indicate the bronchoscope positions achieved by the proposed autonomous navigation and by two expert operators.}
	\label{fig:supp6} 
\end{figure}

\begin{figure} 
	\centering
	\includegraphics[width=\textwidth]{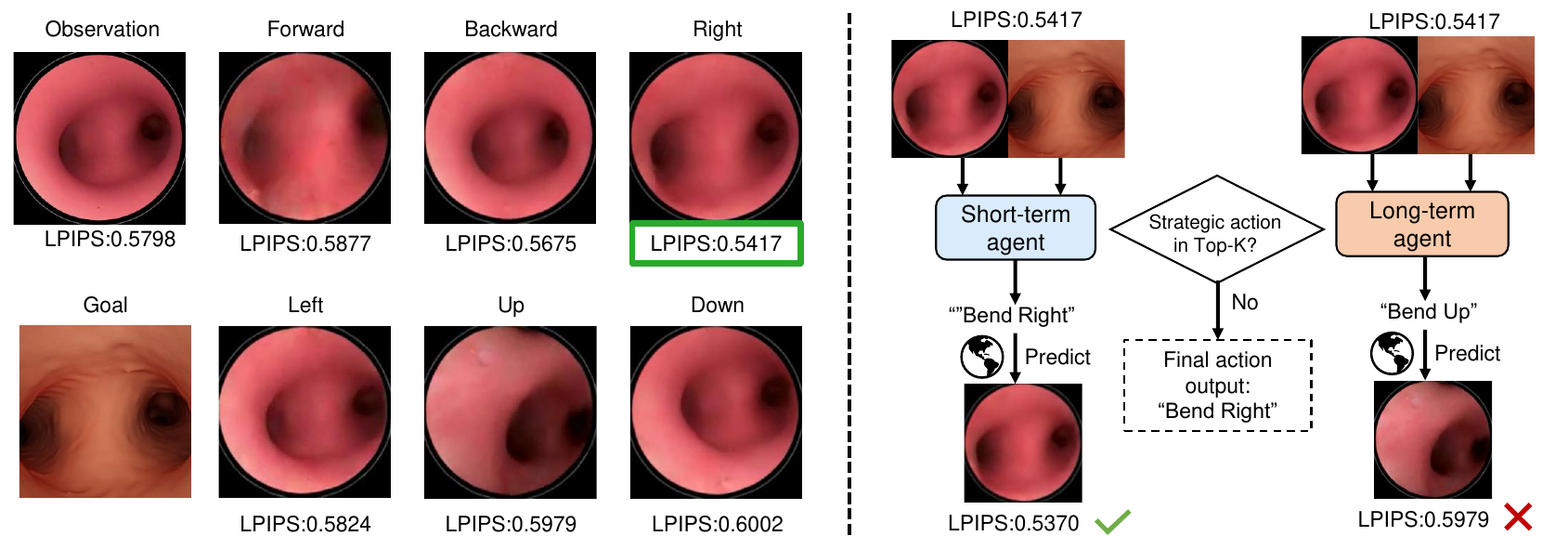} 

	\caption{\textbf{The role of world model in the navigation framework}. (A) Future state predictions generated by the world model under different candidate actions, together with the corresponding LPIPS distances to the goal image. (B) When the short-term agent and long-term agent yield inconsistent action estimates, the world model predicts the resulting future states for each candidate action and evaluates their similarity to the virtual target. The action associated with the highest similarity is selected as the optimal action.}
	\label{fig:supp9} 
\end{figure}

\begin{figure} 
	\centering
	\includegraphics[width=\textwidth]{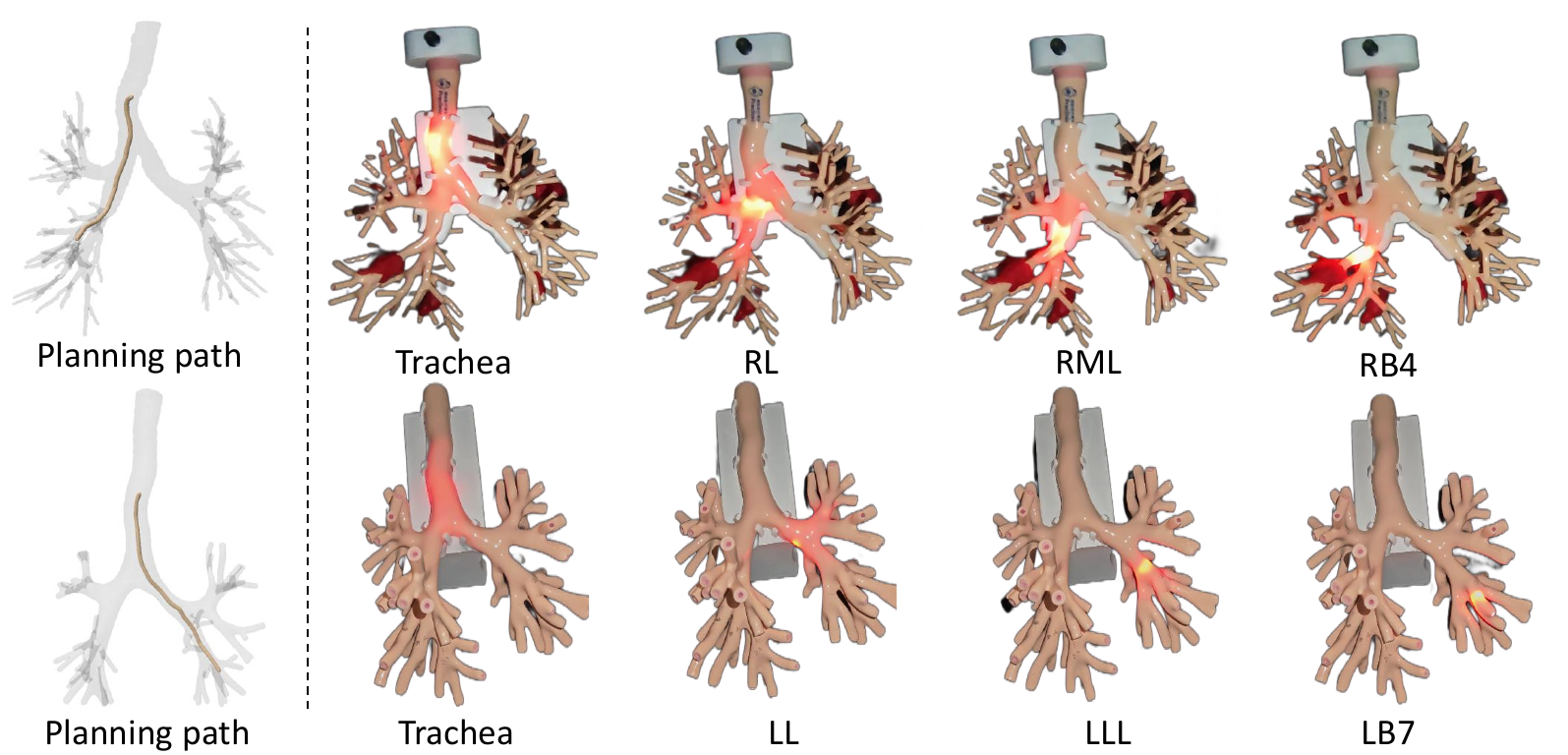} 

	\caption{\textbf{Navigation process in two phantoms.} Illuminated paths indicate the navigation trajectories. Two distinct trajectories are recorded across two different phantoms.}
	\label{fig:supp13} 
\end{figure}

\begin{figure} 
	\centering
	\includegraphics[width=\textwidth]{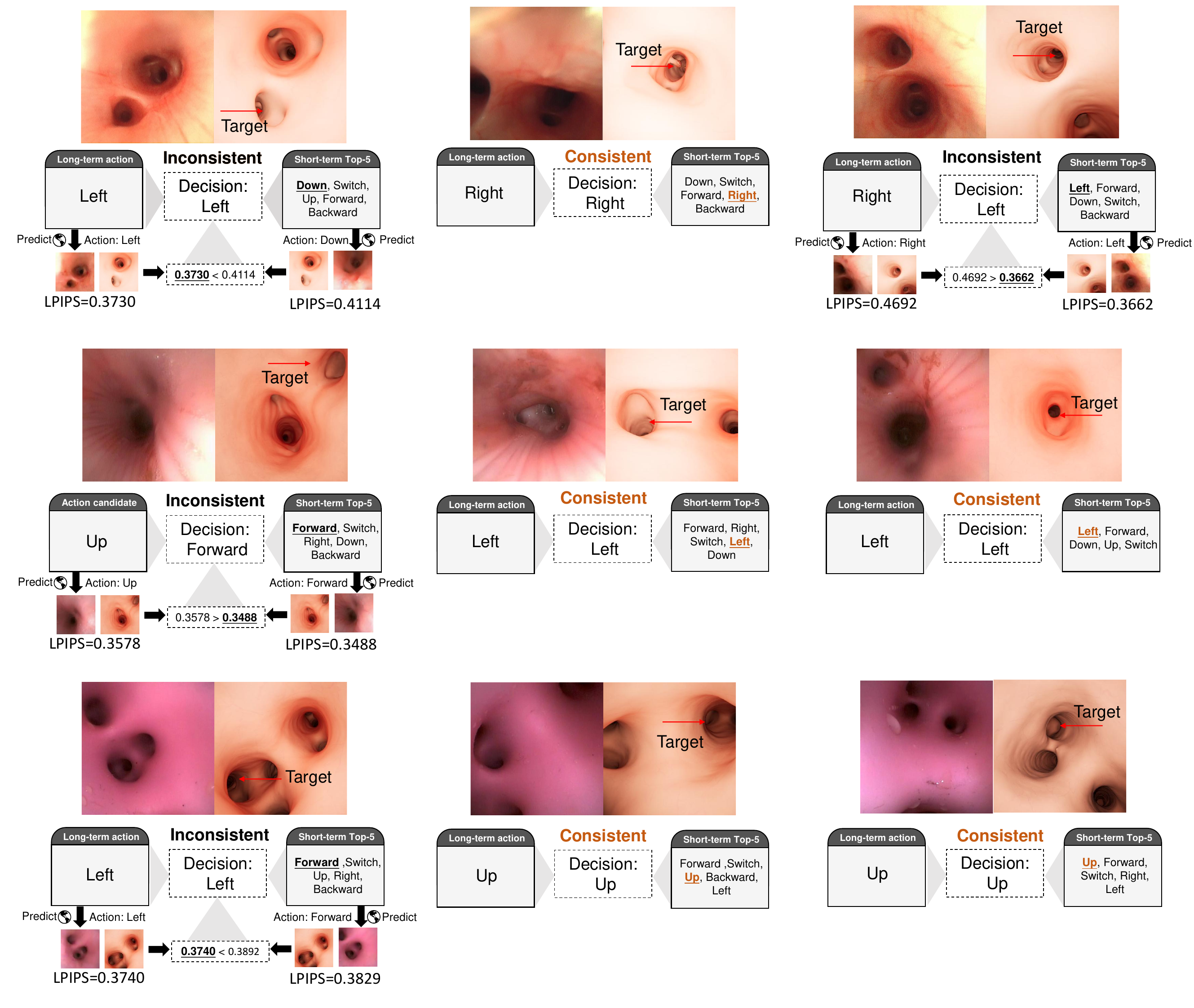} 

	\caption{\textbf{Working mechanism of LLM-based guidance.} The LLM-based guidance framework is illustrated with representative examples from phantom, ex vivo, and in vivo settings. In the virtual view, arrows pointing toward the target lumen provide high-level semantic cues that prompt the LLM to propose an action candidate. During the procedure, the agent evaluates real-time observations against the virtual target and produces probability logits over the action space. If the LLM-proposed action falls within the top-5 logits, it is directly accepted. Otherwise, a world model is invoked to simulate future states under alternative actions, and the action whose predicted future state exhibits the highest similarity to the virtual target is selected as the final decision.}
	\label{fig:supp15} 
\end{figure}

\begin{figure} 
	\centering
	\includegraphics[width=\textwidth]{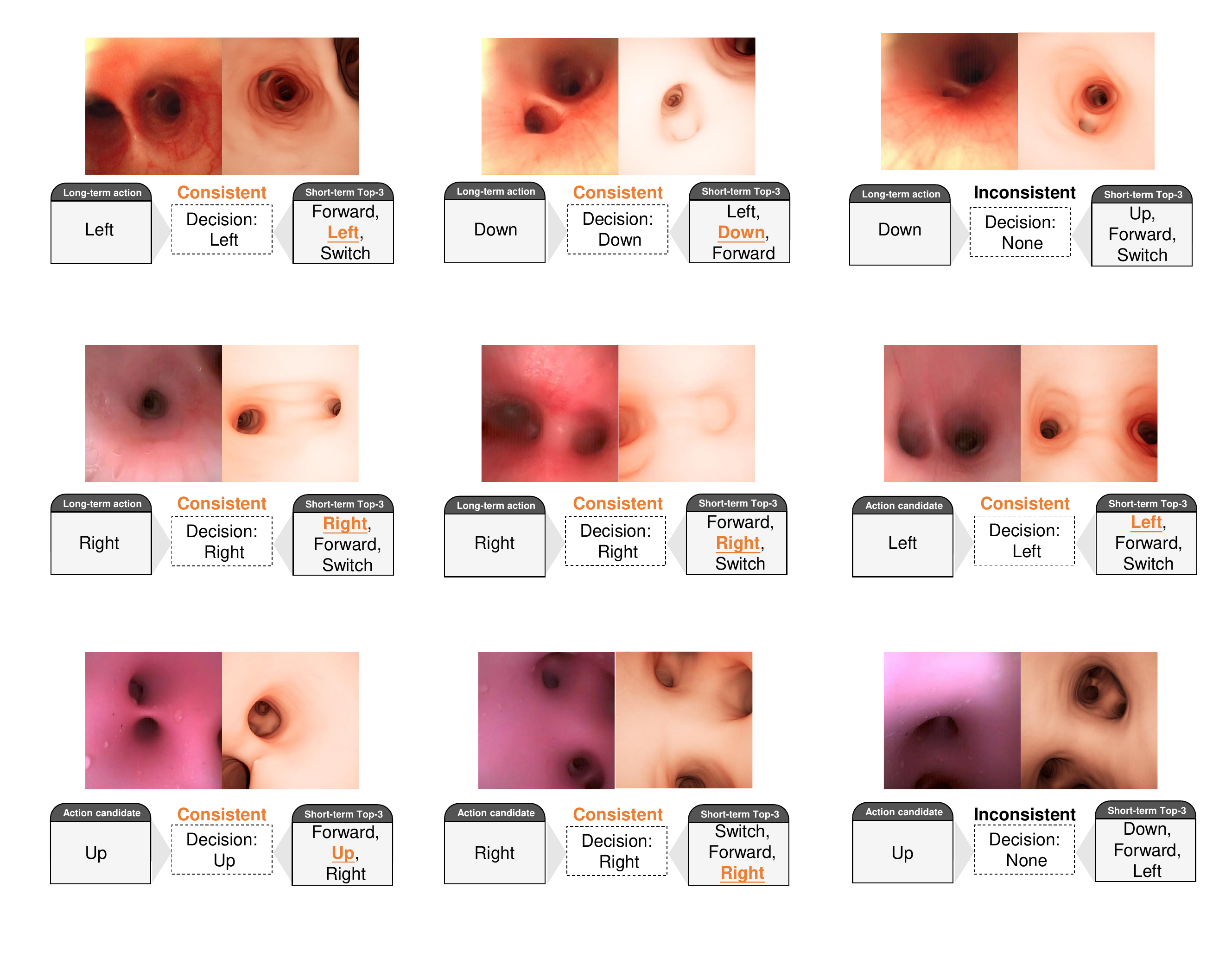} 

	\caption{\textbf{Working mechanism of preoperative guidance.} The preoperative guidance strategy is illustrated with representative examples from phantom, ex vivo, and in vivo settings. Action candidates are derived from preoperative centerline coordinate variations along the planned path. During the intra-operative phase, the agent evaluates each candidate using real-time observations and the corresponding virtual target, producing probability logits over the action space. An action is executed only if the candidate falls within the top-3 logits; otherwise, the system outputs None and no action is performed, ensuring conservative and robust decision-making under uncertainty.}
	\label{fig:supp16} 
\end{figure}

\begin{figure} 
	\centering
	\includegraphics[width=\textwidth]{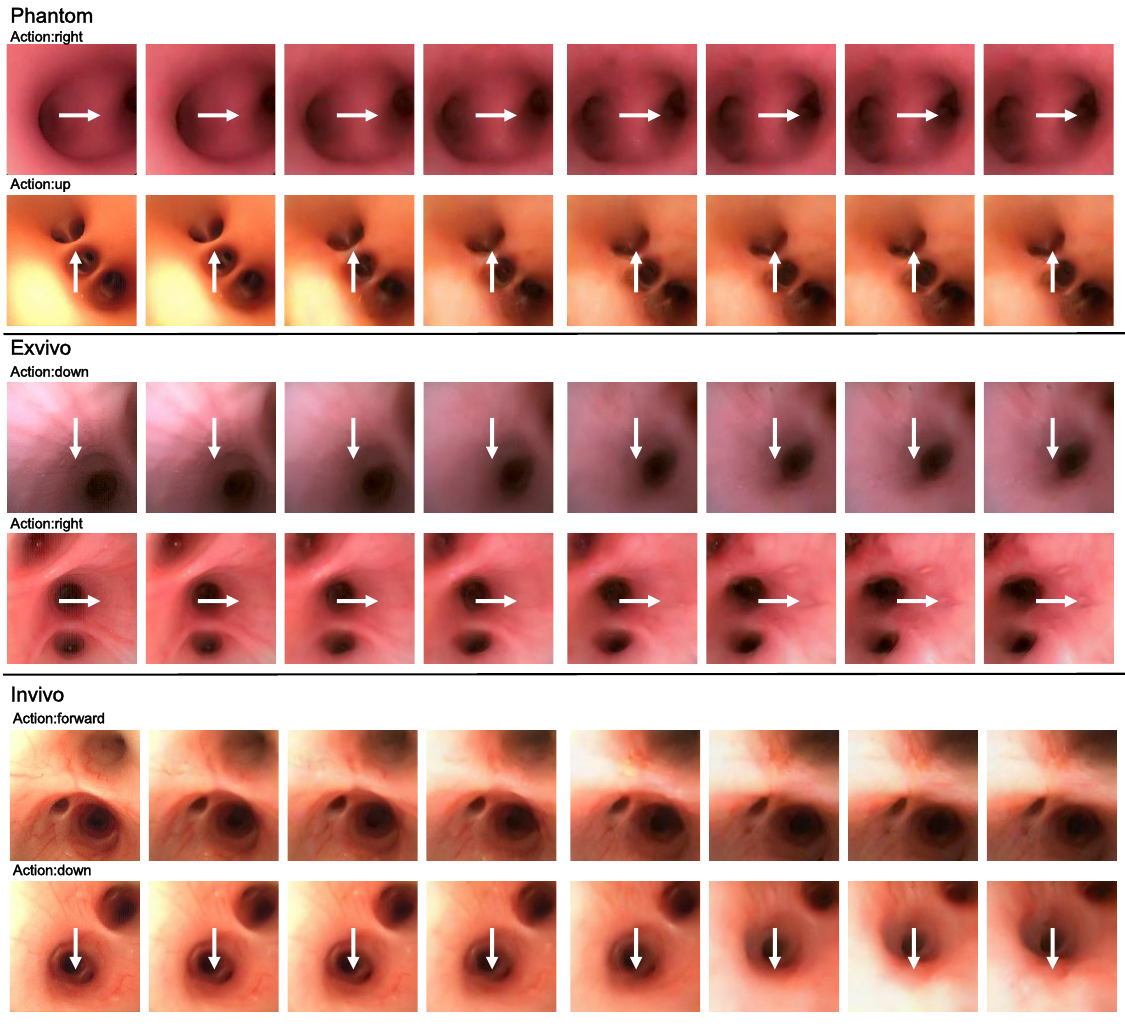} 

	\caption{\textbf{Future prediction of the world model}. The world model receives the first one frame of an unseen video as context input and the predicts next 15 steps into the future given the action sequence, without access to intermediate images}
	\label{fig:supp17} 
\end{figure}

\begin{table}[htbp]
\centering
\caption{\textbf{Timing analysis of the proposed hierarchical agent architecture.}
The system is explicitly decoupled across temporal abstractions. The short-term agent operates continuously for reactive control and low-level decision-making, whereas the long-term agent is event-driven and invoked only at sparse, task-dependent milestones for high-level reasoning and planning.
Trigger frequency and per-call latency are reported as mean $\pm$ standard deviation over $N=5$ independent trials. Reported latencies correspond to algorithmic computation or physical execution time per trigger. For short-term robot execution, a fixed waiting period of 3\,s is enforced after each action before the next control cycle begins, ensuring completion of the current motion; therefore, the robot execution latency is reported as 3\,s per trigger. Decision actions without physical execution (e.g., subgoal switching) are omitted, as they do not include robot execution time.}

\label{tab:timing_analysis}
\begin{tabular}{@{}llcccc@{}}
\toprule
\textbf{Agent Level} & \textbf{Sub-process} & \textbf{Trigger Type} 
& \textbf{Triggers / Trial} 
& \textbf{Per-call Latency (s)} \\
\midrule
\multirow{2}{*}{\textbf{Short-Term}} 
& Model Inference      & Continuous   & 76.6 $\pm$ 20.206 & 0.0065 $\pm$ 0.0012 \\
& Robot Execution      & Continuous   & 76.6 $\pm$ 20.206 & 3 $\pm$ 0\\
\midrule
\multirow{2}{*}{\textbf{Long-Term}} 
& Preoperative Guidance & Event-driven & 22.6 $\pm$ 5.177 & 0.715 $\pm$ 0.502 \\
& VLM Reasoning     & Event-driven & 1  $\pm$ 0  & 42.339 $\pm$ 7.773  \\
\bottomrule
\end{tabular}
\end{table}



\begin{table}[t]
\centering
\caption{Core Robot Configuration}
\label{tab:core_robot_config}
\setlength{\tabcolsep}{8pt}
\renewcommand{\arraystretch}{1.12}
\begin{tabular}{@{}p{0.62\linewidth}p{0.24\linewidth}@{}}
\toprule
\rowcolor{gray!20}
\textbf{Parameter} & \textbf{Value} \\
\midrule
default translation speed (m/s) & 0.5 \\
default rotation speed (deg/s) & 20 \\
default bending speed (deg/s) & 10 \\
bending range (deg) & [-210, 130] \\
rotation step (deg) & 5 \\
bending step (deg) & 9 \\
translation step (m) & 0.002 \\
\bottomrule
\end{tabular}
\end{table}








\end{document}